\documentclass{styles/egpubl}

\ConferencePaper        %

\usepackage[T1]{fontenc}
\usepackage{styles/dfadobe}

\usepackage{cite}  %
\BibtexOrBiblatex
\electronicVersion
\PrintedOrElectronic
\ifpdf \usepackage[pdftex]{graphicx} \pdfcompresslevel=9
\else \usepackage[dvips]{graphicx} \fi

\usepackage{styles/egweblnk}

\usepackage{graphicx, calc}
\usepackage{multirow}
\usepackage{makecell}
\usepackage{colortbl}
\usepackage{xcolor}
\usepackage{xspace}
\usepackage{amsmath}
\usepackage{booktabs}
\usepackage{amsfonts}
\usepackage{url}
\usepackage{bbold}
\usepackage[normalem]{ulem}

\definecolor{rred}{RGB}{216, 27, 96}
\definecolor{rblue}{RGB}{30, 136, 229}
\definecolor{rgreen}{RGB}{28, 151, 77}
\definecolor{ryellow}{RGB}{255, 193, 7}
\definecolor{rorange}{RGB}{255,167,0}
\definecolor{americanrose}{rgb}{1.0, 0.01, 0.24}
\definecolor{cadmiumgreen}{rgb}{0.0, 0.42, 0.24}
\definecolor{ryb}{rgb}{0.1, 0.29, 0.8}
\definecolor{cc1}{HTML}{ee6352}
\definecolor{cc2}{HTML}{59cd90}
\definecolor{cc3}{HTML}{3fa7d6}
\definecolor{cc4}{HTML}{fac05e}
\definecolor{cc5}{HTML}{f7a9a8}
\definecolor{cc6}{HTML}{77b05f}
\definecolor{cc7}{HTML}{a64d79}
\definecolor{cc8}{HTML}{3c78d8}
\definecolor{c1}{HTML}{2196f3}
\definecolor{c2}{HTML}{4caf50}
\definecolor{c3}{HTML}{ff9800}
\definecolor{c4}{HTML}{f44336}
\definecolor{c5}{HTML}{673ab7}
\definecolor{originalcolor}{HTML}{D79B00}
\definecolor{transformedcolor}{HTML}{82B366}

\newif\ifEDITINGFLAG
\newcommand{\IL}[1]{\ifEDITINGFLAG\textcolor{cadmiumgreen}{#1}\else #1\fi}
\newcommand{\JF}[1]{\ifEDITINGFLAG\textcolor{ryellow}{#1}\else #1\fi}
\newcommand{\RC}[1]{\ifEDITINGFLAG\textcolor{rblue}{#1}\else #1\fi}

\newcommand{\datasetsize}{eight}
\newcommand{\figurecount}{eight}
\newcommand{\scene}[1]{s_#1}

\newcommand{\Szero}{s_0}
\newcommand{\Sone}{s_1}

\newcommand{\s}[1]{s^{\text{\romannumeral #1}}}

\newcommand{\Tk}[1]{T_#1}
\newcommand{\T}{T}
\newcommand{\F}{\mathcal{F}}
\newcommand{\Fi}[1]{\mathcal{F}_{#1}}
\newcommand{\Cdiff}[1]{{#1}}
\newcommand{\Cspec}[1]{{#1}}

\newcommand{\albedo}{\rho}
\newcommand{\rough}{r}

\newcolumntype{C}[1]{>{\centering\let\newline\\\arraybackslash\hspace{0pt}}m{#1}}
\newcolumntype{S}[1]{@{\hspace{#1}}}
\newcolumntype{H}{>{\setbox0=\hbox\bgroup}c<{\egroup}@{}}

\newcommand{\verti}[2]{\multirow{1}{*}[#1]{\rotatebox[origin=c]{90}{#2}}}
\newcommand{\oc}[1]{\cellcolor{gray!10}}

\newcommand{\newparagraph}[1]{\vspace{0.4em}\noindent\textbf{#1}\quad}

\DeclareFontFamily{U}{mathb}{}
\DeclareFontShape{U}{mathb}{m}{n}{
  <-5.5> mathb5
  <5.5-6.5> mathb6
  <6.5-7.5> mathb7
  <7.5-8.5> mathb8
  <8.5-9.5> mathb9
  <9.5-11.5> mathb10
  <11.5-> mathb12
}{}
\DeclareSymbolFont{mathb}{U}{mathb}{m}{n}
\DeclareMathSymbol{\ulsh}{3}{mathb}{"E8}
\DeclareMathSymbol{\ursh}{3}{mathb}{"E9}
\DeclareMathSymbol{\dlsh}{3}{mathb}{"EA}
\DeclareMathSymbol{\drsh}{3}{mathb}{"EB}

\makeatletter
\DeclareRobustCommand\onedot{\futurelet\@let@token\@onedot}
\def\@onedot{\ifx\@let@token.\else.\null\fi\xspace}

\def\eg{\emph{e.g}\onedot} 
\def\ie{\emph{i.e}\onedot} 
 
\def\etc{\emph{etc}\onedot} 
\def\wrt{w.r.t\onedot} 

\makeatother

\usepackage[capitalize]{cleveref}
\crefname{section}{sec.}{secs.}
\Crefname{section}{Sec.}{Secs.}
\crefname{paragraph}{sec.}{secs.}
\Crefname{paragraph}{Sec.}{Secs.}
\crefname{table}{Tab.}{Tabs.}
\Crefname{table}{Tab.}{Tabs.}
\crefname{figure}{Fig.}{Figs.}
\Crefname{figure}{Fig.}{Figs.}
\crefname{equation}{eq.}{eqs.}
\Crefname{equation}{Eq.}{Eqs.}

\title[Material transforms from disentangled NeRF representations]%
      {Material transforms from disentangled NeRF representations}
\author[I. Lopes \& JF. Lalonde \& R. de Charette]
{\parbox{\textwidth}{\centering Ivan Lopes$^{1}$ \quad Jean-François Lalonde$^{2}$ \quad Raoul de Charette$^{1}$ \\[1ex] \centering $^1$Inria \, $^2$Université Laval}}

\begin{document}

\teaser{
 \includegraphics[width=.95\linewidth]{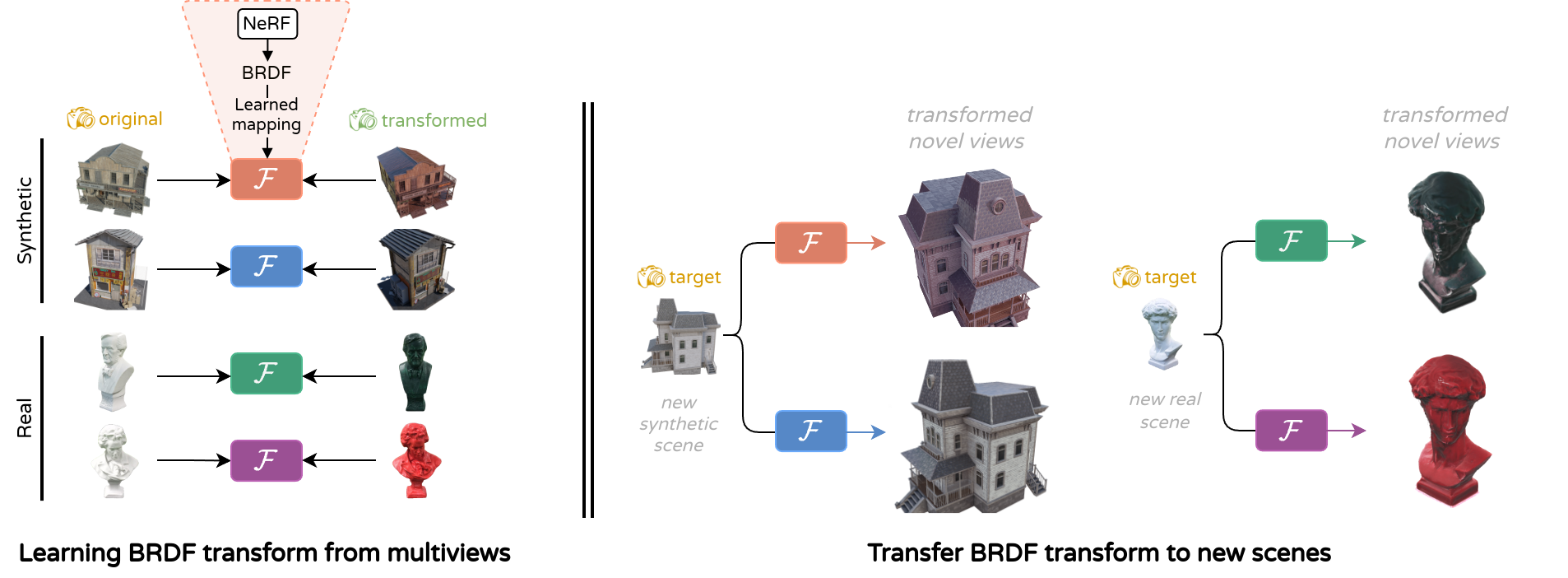}
 \centering
  \caption{\textbf{Proposed method.} We illustrate our approach for inferring unknown material transformations in complex scenes. From a set of observations of a scene in two conditions: {\color{originalcolor}original} and {\color{transformedcolor}transformed}, we leverage a joint Neural Radiance Field (NeRF) optimization to learn a material mapping function $\mathcal{F}$ which models the observed changes at the material level accurately (\eg the topmost transform on the left is a red varnish). This learned function can be applied to new {\color{originalcolor}target} scenes with different geometry and material properties (right).
 }
\label{fig:teaser}
}

\maketitle
\begin{abstract}
In this paper, we first propose a novel method for transferring material transformations across different scenes. Building on disentangled Neural Radiance Field (NeRF) representations, our approach learns to map Bidirectional Reflectance Distribution Functions (BRDF) from pairs of scenes observed in varying conditions, such as dry and wet.
The learned transformations can then be applied to unseen scenes with similar materials, therefore effectively rendering the transformation learned with an arbitrary level of intensity.
Extensive experiments on synthetic scenes and real-world objects validate the effectiveness of our approach, showing that it can learn various transformations such as wetness, painting, coating, etc. Our results highlight not only the versatility of our method but also its potential for practical applications in computer graphics. We publish our method implementation, along with our synthetic/real datasets on \url{https://github.com/astra-vision/BRDFTransform}

\printccsdesc
\end{abstract}

\section{Introduction}
\label{sec:intro}

In computer graphics and vision, inverse rendering is key to extracting material information and allowing re-rendering under novel conditions (viewpoint, lighting, materials, etc.). While neural representations have largely taken over the traditional Physically-Based Rendering~(PBR) techniques, recent works have demonstrated that the two representations can be combined~\cite{jin2023tensoir}, thus preserving the editability and expressivity of PBR representations along with the flexibility of neural representations. 

When considering the appearance of a scene, certain 
\RC{transformations (such as applying a coat of varnish)}
can alter the material properties significantly, causing the scene's appearance to change drastically. Currently, estimating the PBR characteristics of a known material after such a transformation requires capturing the scene again in the desired target condition. This process is both complex and laborious due to the variety of possible transformations, such as wetness, dust, varnish, painting, \etc.
In this work, we aim to learn a BRDF transformation from a source scene \IL{and apply it to different scenes.}

Assuming we have paired observations of the same scene under two different conditions, say \textit{original} and~\textit{varnished}, we propose a method to learn the transformation of materials. This transformation can then be applied to another scene composed of similar materials. This allows us to predict the appearance of that scene under this effect, effectively transferring the material transformation.

\RC{\cref{fig:teaser} illustrates that several material transformations can be learned from multiple pairs of scenes (left) and later applied on novel scenes (right), whether synthetic or real.
Technically, our method} relies on the joint optimization of \RC{a radiance field}
corresponding to a first scene captured in original and transformed (\eg, varnished) conditions, possibly with varying lighting conditions.
We rely here on the disentangled NeRF representation of \mbox{TensoIR}~\cite{jin2023tensoir}---that optimizes appearance, geometry, and parametric BRDF simultaneously---while introducing two novel key components. 
First, we condition the transformed scene BRDF on the original scene and approximate its transformation with \IL{a Multi-Layer Perceptron (MLP)}. 
Second, we expose a limitation of \mbox{TensoIR} showing \IL{it} \RC{fails} %
at decomposing highly reflective materials and propose an improved light estimation scheme that better estimates low roughness components while preserving high frequencies in the illumination.
As a result, our framework allows capturing a collection of transformations which can then be applied on new scenes, while controlling the intensity of the \RC{transformation}.
\RC{We demonstrate the performance of our method on two new datasets: a synthetic dataset with a series of custom shader transformations and a real-world dataset of figurines with varying material conditions (\eg, original, painted, varnished, \etc). On both datasets, our approach produces faithful transformations.}
Our method \RC{and datasets} will be released publicly.

\section{Related work}
\label{sec:related}

Inverse rendering is a long-standing problem, it has gained interest recently with the use of neural radiance fields \cite{mildenhall2020nerf}. Given a set of images of an object taken from different points of view, the goal is to optimize an implicit volumetric model for opacity and radiance. This allows synthesizing frames at novel viewpoints using volume rendering. Many works have extended this approach to learning a more explicit volume in which material information is disentangled from light sources. This way, the scene can be relit and the material manipulated, providing much more control over conventional radiance-centered methods. 

\begin{figure*}[ht!]
    \centering
    \includegraphics[width=.99\linewidth]{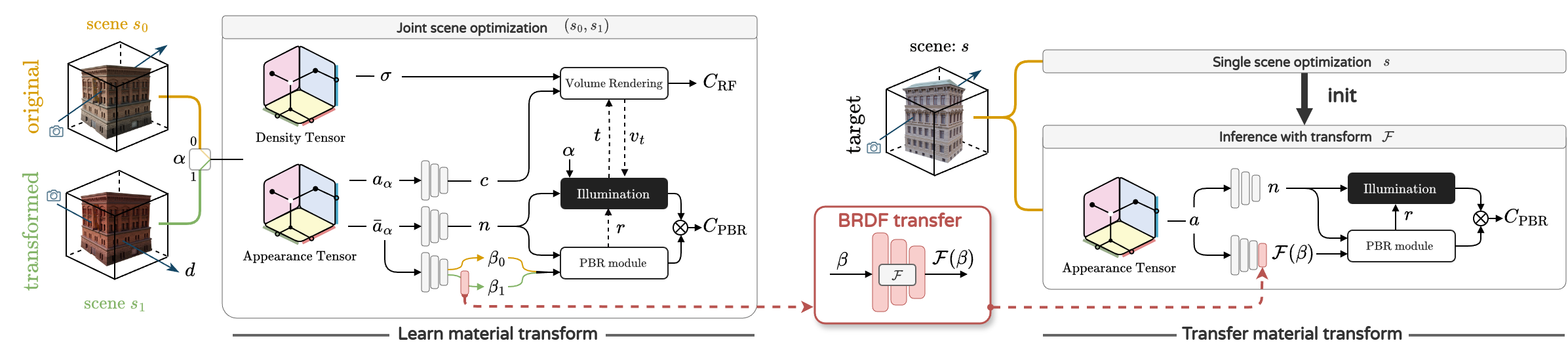}
    \caption{\textbf{Overview of our proposed method.} Our method takes observations of the same scene with two different materials \IL{$(\beta_0, \beta_1)$ for $\Szero$ and $\Sone$, respectively. We assume $\beta_1$ to be a function of $\beta_0$.}
    Our method learns a joint~representation and a transform function $\Fi{}$ which maps the material of the first to the second (left block). Given a new scene $s$, we learn its geometry and material and apply our learned transform function (right block) to produce the same effects observed in the source scenes $(\Szero, \Sone)$.
    }
    \label{fig:main}
\end{figure*}

\newparagraph{BRDF estimation in NeRF.} NeRD \cite{boss2021nerd} is the first method to perform BRDF optimization of a scene in an uncontrolled setting. Later, approaches such as NeRV~\cite{srinivasan2021nerv} and IndiSG~\cite{zhang2022modeling} introduced solutions for self-occlusions and indirect light. Spherical Gaussians (SG) have been widely used for modeling illumination in inverse rendering~\cite{physg2021,zhang2022modeling,jin2023tensoir,zhang2023nemf}. Implicit representations were subsequently introduced in NeILF/++~\cite{yao2022neilf,zhang2023neilf++}, NeRO~\cite{liu2023nero} and TensoSDF~\cite{li2024tensosdf} to better represent high frequency illumination. For specular objects, some have proposed new forms of encodings to help supervise narrow specular lobes. For example, Ref-NeRF~\cite{verbin2022refnerf} uses the Integrated Directional Encoding (IDE), NeAI~\cite{zhuang2023neai} an Integrated Lobe Encoding (ILE), and SpecNeRF~\cite{ma2023specnerf} Gaussian directional encodings. 
These optimization methods have been combined with Signed Distance Functions (SDF) in Factored-NeuS~\cite{fan2023factoredneus} and NeRO~\cite{liu2023nero} to provide a more robust geometry estimation. Recently, NeP~\cite{wang2024inverse} uses a neural plenoptic function to model incoming light.
Unlike others who adopted analytical BRDFs, NeRFactor \cite{Zhang2021nerfactor} uses a data-driven approach by first learning priors on real-world BRDFs from the MERL dataset~\cite{Matusik2003merl}. Instead, ENVIDR~\cite{liang2023envidr} learns this prior on a synthetic dataset.
NVDiffrec/-MC \cite{Munkberg2022NVDiffrec,hasselgren2022nvdiffrecmc} optimize the mesh and its materials as SVBRDF maps. 

Closest to our approach in terms of scene optimization, TensoIR~\cite{jin2023tensoir} adopts a tensor representation and factorizes a light component to learn under multiple illumination. They use stratified sampling and SGs to model direct light, while we adopt a neural representation. NeRO~\cite{liu2023nero} has the same illumination approach but uses a two-stage approach which is computationally expensive. \IL{Instead, we use an approximation of the rendering equation to pre-compute part of the integral.}

\newparagraph{Material and neural transforms.} While the problem of BRDF \IL{transform} in a multi-view setting has not been explored, to the best of our knowledge, we present relevant research on this topic. In tvBRDF~\cite{sun2007time}, the authors propose analytical models for transforms such as dust, watercolors, oils, and sprays, \IL{on non-spatially varying materials}. Another line of research looks at translating a NeRF reconstruction based on an exemplar-style image. This includes StyleNeRF~\cite{liu2023stylerf}, LAENeRF~\cite{radl2023laenerf}, or iNeRF2NeRF~\cite{instructnerf2023} which is prompt-based. Also related is the task of performing material transfers such as in NeRF-analogies~\cite{fischer2024nerf}. In Climate-NeRF~\cite{Li2023ClimateNeRF}, global effects are injected into the scene but do not affect the object materials.

\newparagraph{Inverse rendering datasets.} Datasets with varying BRDFs were introduced in \cite{gu2006time} with time-varying effects. They record a number of surfaces transformed by natural processes showing how it affects the BRDF temporally and spatially. It is common that inverse rendering datasets offer captures under different illuminations but the material remains unchanged: ReNe~\cite{toschi2023relight} proposes a dataset of 20 real scenes captured under 40 point-light positions. Objects-with-Lighting~\cite{ummenhofer2024objects} introduces 8 objects under 3 environments with the corresponding \IL{High Dynamic Range (HDR) environment maps}.

\section{Method}
\label{sec:method}

\setlength{\abovedisplayskip}{3pt}
\setlength{\belowdisplayskip}{3pt}

\subsection{Problem setting}
\label{sec:problem}

Consider the scenario shown in \cref{fig:main}, where a scene is observed in its \textit{original} state $s_0$, and a second time \IL{$s_1$ with its materials \emph{transformed} by an unknown effect $T$, such that $s_1{} = T(s_0)$.}
For~example, $T$ could be the result of applying \IL{a coat of paint, some colored varnish, or having the scene soaked with water}.
Note that~$\Sone$ might have been captured under a different illumination than~$\Szero$. 
Our goal is to model the material transformation happening between~\IL{$\Szero$~and~$\Sone$}, in such a way that we can transfer this effect \IL{to a new scene.}

We model the scene with a BRDF field, more specifically, every point \IL{of a scene $s$} is characterized by material properties ${\beta = (\albedo,\rough) \in \mathbb{R}^4}$, where $\albedo$ is the albedo (in RGB) and $\rough$ the roughness. \IL{\RC{Our formulation assumes that the original} scene $s_0{}$ is affected by an unknown transformation which changes its material properties $\beta_0$ \RC{but not its geometry, resulting in scene} $s_1$ with material $\beta_1$. Our method aims to learn a function $\Fi{}{}$ which approximates the unknown mapping $\T$ between the two materials $\beta_0$ and $\beta_1$}, in such a way that $\Fi{}{}$ can be applied on new scenes as shown~in~\cref{fig:main}~\IL{(\textit{right})}.

\subsection{Preliminaries}

Our optimization approach is based on TensoIR \cite{jin2023tensoir}, itself derived from TensoRF~\cite{chen2022tensorf}, to learn a neural radiance field of the scene. For clarity, we follow their notation here. In this framework, a \IL{radiance field is learned by} jointly training both a density tensor $\mathcal{G}_{\sigma}$ and an appearance tensor $\mathcal{G}_{a}$. From the latter, surface normals $n$ and material properties $\beta$ can be estimated at every 3D point $x$ using lightweight MLPs, \IL{noted $\mathcal{D}$, and accumulated along each viewing rays} using volume rendering. While the scene can be imaged under a single illumination condition, TensoIR also supports multiple observations of the scene under different illuminations. In this case, it further factorizes a light embedding to produce light-dependent appearance features $a_{\alpha}$, where $\alpha$ indexes the lighting conditions (modeled as an environment map).
The estimated quantities at every point $x$ of the scene can therefore be written as: 
\begin{equation}
    n = \mathcal{D}_{n}(\IL{\bar{a}_\alpha}),\quad \beta = \mathcal{D}_{\beta}(\IL{\bar{a}_\alpha}) ,\quad c_\alpha = \mathcal{D}_{c}(a_\alpha)\,,
\end{equation}
where \IL{$\bar{a}_\alpha$} is the average appearance features across both light embeddings, and $c$ is the pixel color (as in the original TensoRF formulation). 
\IL{TensoIR learns a disentangled representation, allowing the color of each point \( x \) to be estimated for a given view direction \( d \) either through volume rendering, represented as $C_{\text{RF}}(x, d)$, or through physically-based rendering, also represented as $C_{\text{PBR}}(x, d)$ — both of which are supervised by the reference images.}

\subsection{Learning material transforms}
\label{sec:meth_mattransform}

As discussed in \cref{sec:problem}, we aim to learn $\F{}$ which maps the BRDF parameters $\beta_0$ of a scene $s_0$ to its transformed appearance $\beta_1$ for $s_1$. As illustrated in \cref{fig:main}, we formulate the transfer with:%
\begin{equation}
    \IL{\beta_\alpha=\beta_0[\alpha=0] + \F{}(\beta_0)[\alpha=1]} 
    \label{eq:mapping}
\end{equation}
\IL{where $[\bullet]$ is the Iverson bracket and} $\F{}$ is a small MLP network that is trained end-to-end together with the \IL{appearance and density tensors}. \IL{Here $\alpha$ is an indicator} representing whether we are rendering the original scene \RC{(\ie $\alpha=0$)} or its transformed version \RC{(\ie $\alpha=1$)}. \IL{Using this formulation, we jointly train on $s_0$ and $s_1$, and learn a single neural representation for both scenes.}

\subsection{Light estimation}

\begin{figure}[!h]
    \centering
    \renewcommand{\arraystretch}{0.3}
    \setlength{\tabcolsep}{1pt} %
    \newcommand{\imgW}{.5\linewidth}
    \newcommand{\imgH}{.25\linewidth}

    \newcommand{\row}[1]{%
        \includegraphics[height=\imgH]{images/toy/#1_render.png}&%
        \includegraphics[height=\imgH]{images/toy/#1_rough.png}&%
        \includegraphics[height=\imgH]{images/toy/#1_envmap.png}%
    }
    \resizebox{.99\linewidth}{!}{%
    \begin{tabular}{C{3ex} cS{.5ex}cS{.5ex}c}
        & Render & Roughness & Envmap \\

         \verti{10ex}{GT} & \row{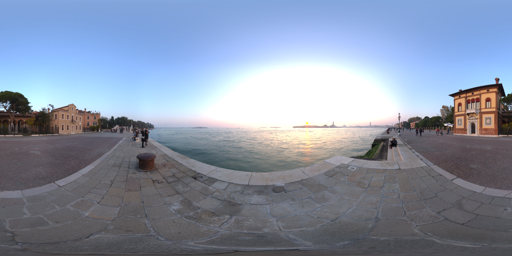}\\
         \verti{10ex}{TensoIR} & \row{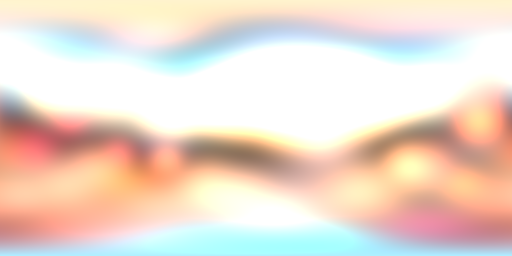}\\

    \end{tabular}}
    
    \caption{\textbf{TensoIR \IL{on glossy surfaces}.} We \IL{observe that TensoIR overestimates roughness and smoothes the estimated illumination}.}
    \label{fig:tensoir-roughness-demo}
\end{figure}

\newparagraph{\IL{Limitation of TensoIR}.} We observe that the original TensoIR framework struggles in reconstructing low-roughness scenes~(\cref{fig:tensoir-roughness-demo}), \IL{which is crucial for representing glossy surfaces}. We also note \RC{that} the low number of spherical gaussians used to represent the environment results in the absence of high-frequency content in the lighting. The use of stratified sampling and low-frequency light representation comes at the cost of incorrect estimation of objects with low roughness.\\
\RC{To alleviate this problem} and allow learning a wider variety of material transforms, we propose an improvement to the formulation by borrowing ideas from NeRO~\cite{liu2023nero}. We keep the volume representation of TensoIR as it is fast to optimize but avoid expensive light sampling by following NeRO. \IL{That way, we benefit from both methods and ensure fast optimization speeds}.

\newparagraph{Formulation.}
Rendering the color of a point $x$ from a viewing direction $d$ is given by
\begin{equation}
    \IL{C_{\text{PBR}}}(x,d)=\int_{\Omega}{L(\omega,x)f_r(\omega,d)(\omega \cdot n)d\omega} \,,
\end{equation}
where $\Omega$ is the integrating hemisphere, $L$ the light intensity from direction $\omega$ at $x$. Here, the BRDF $f_r$ is parameterized with material properties $\beta=\IL{(\rho,r)}$.  We adopt the micro-facet reflectance model of \cite{cook1982reflectance}: 
\begin{equation}
    f_r(\omega,d) = \Cdiff{\frac{\albedo}{\pi}} + \Cspec{\frac{DFG}{4(\omega \cdot n)(d \cdot n)}} \,,
\end{equation}
where $D$, $F$, and $G$ are the normal distribution, Fresnel, and geometric attenuation terms. \IL{For brevity, we omit the parameters for these three functions.} 
We follow NeRO~\cite{liu2023nero} and use the split-sum approximation on the specular component~\cite{Karis2013RealSI}. After integration, it becomes:
\begin{equation}
    \IL{C_{\text{PBR}}}(x,d) = \Cdiff{\albedo \ell_\mathrm{diff}} + \Cspec{M_\mathrm{spec} \ell_\mathrm{spec}} \,,
    \label{eq:pb-simple}
\end{equation}
where
\begin{equation}
\Cdiff{\ell_\mathrm{diff} = \int_{\Omega} L(\omega,x)D(n,1) d\omega} \,, \quad 
\Cspec{\ell_\mathrm{spec} = \int_{\Omega} L(\omega,x)D(t,\rough) d\omega} \,,
\end{equation}
and 
\begin{equation}
\Cspec{M_\mathrm{spec} = \int_{\Omega} \frac{DFG}{4(d\cdot n)} d\omega} \,.
\end{equation}
Here, $t$ is the reflected direction \wrt surface normal $n$. 
Note that $M_\text{spec}$ can be precomputed as it does not depend on $L$. The integrals $\ell_\mathrm{diff}$ and $\ell_\mathrm{spec}$ (which depend on $L$) are discussed next.

\begin{figure}
    \centering
    \includegraphics[width=\linewidth,trim={0 .3cm 0 0},clip]{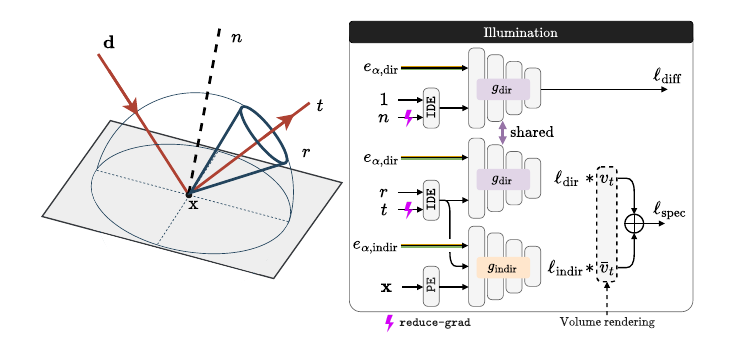}
    \caption{\textbf{Light estimation.} We adopt a neural light representation \cite{liu2023nero} which models direct and indirect light sources separately. On the indirect component, the two types of light sources are blended using an occlusion mask obtained via secondary ray casting along reflected light direction $t$ \cite{jin2023tensoir}. To avoid disrupting the optimization of the geometry, we reduce the gradient intensity along the directional inputs (on $n$ and $t$). \IL{We note $\bar{v}_t = 1 - v_t$, \texttt{IDE} is an Integrated Directional Encoding \cite{verbin2022refnerf} while \texttt{PE} is a Positional Encoding \cite{mildenhall2020nerf}.}
    }
    \label{fig:light}
\end{figure}
\newparagraph{Light estimation.} 
We use the Integrated Directional Encoding~(IDE) of Ref-NeRF \cite{verbin2022refnerf} to model the scene illumination. 
Similar to NeRO~\cite{liu2023nero}, we leverage two MLPs for approximating $L$\RC{, being} $g_{\text{dir}}$ for direct and $g_{\text{indir}}$ for indirect (e.g., interreflections) light. 
To \IL{accommodate for a joint optimization setting on two scenes}, we feed latent embeddings to both light MLPs $g$ in order to account for possible changes \IL{in lighting.}
This is achieved by channel-wise concatenating \IL{the corresponding} embedding to the IDE of the $g$ inputs, depending on whether the original or transformed scene is rendered. The illumination expressions are written as:
\begin{align}
    \begin{aligned}
    \Cdiff{\ell_{\text{diff}} =\ } &\Cdiff{g_{\text{dir}}\left(\texttt{IDE}(n,1),\IL{e_{\alpha,\text{dir}}}\right) }, \\
    \Cspec{\ell_{\text{spec}} =\ } &\Cspec{v_{t}g_{\text{dir}}\left(\texttt{IDE}(t,\rough),\IL{e_{\alpha,\text{dir}}}\right) } \\
        &\Cspec{+ (1-v_t)g_{\text{indir}}\left(\texttt{IDE}(t,\rough),x,\IL{e_{\alpha,\text{indir}}}\right) } \,.
    \end{aligned}
\end{align}

For the specular term $\ell_{\text{spec}}$, $v_t$ is the visibility term obtained by ray tracing along $t$ in the density volume. 
A detail of the proposed illumination components is shown in \cref{fig:light}. 
Here, the latent embeddings $e$ provide conditioning to the $g$ networks in order to best model scene-specific illuminations.

\section{Experiments}
\label{sec:exp}

\subsection{Experimental methodology}

\begin{figure}[t]
    \centering
    \scriptsize
    \newcommand{\imgsize}{.123\linewidth}
    \setlength{\tabcolsep}{.2pt} %
    \renewcommand{\arraystretch}{0.5} %

    \newcommand{\row}[3]{%
        \verti{#2}{#1}&%
        \includegraphics[width=\imgsize,height=\imgsize]{images/dataset/fastfood_#3.png}&%
        \includegraphics[width=\imgsize,height=\imgsize]{images/dataset/historical_#3.png}&%
        \includegraphics[width=\imgsize,height=\imgsize]{images/dataset/mansion_#3.png}&%
        \includegraphics[width=\imgsize,height=\imgsize]{images/dataset/motel_#3.png}&%
        \includegraphics[width=\imgsize,height=\imgsize]{images/dataset/station_#3.png}&%
        \includegraphics[width=\imgsize,height=\imgsize]{images/dataset/vintage_#3.png}&%
        \includegraphics[width=\imgsize,height=\imgsize]{images/dataset/store_#3.png}&%
        \includegraphics[width=\imgsize,height=\imgsize]{images/dataset/wildwest_#3.png}}%

    \resizebox{.99\linewidth}{!}{%
    \begin{tabular}{c cccccccc}
        & $\s{1}$ & $\s{2}$ & $\s{3}$ & $\s{4}$ & $\s{5}$ & $\s{6}$ & $\s{7}$ & $\s{8}$ \\
        \row{original}{8ex}{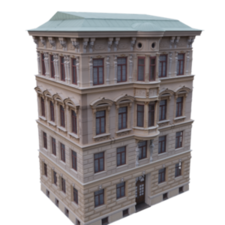}\\
        \row{$\Tk{1}$}{6ex}{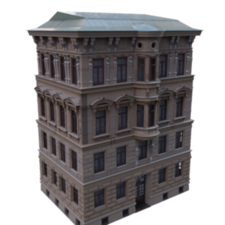}\\
        \row{$\Tk{2}$}{6ex}{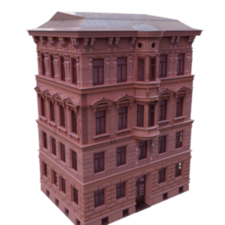}\\
        \row{$\Tk{3}$}{6ex}{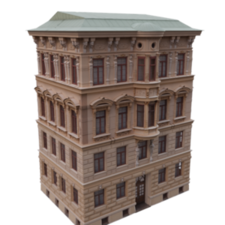}\\
        \row{$\Tk{4}$}{6ex}{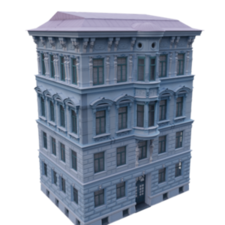}\\
    \end{tabular}}

    \caption{\textbf{\RC{Synthetic dataset.}} Each column shows a difference scene $s^k, k \in \{\text{\romannumeral 1}, \ldots, \text{\romannumeral 8}\}$. The first row shows the original scene, each subsequent row shows the scene after each synthetic transformation $T_j, j \in \{1, \ldots, 4\}$.}
    \label{fig:syn-dataset}
\end{figure}

\begin{figure}[t]
    \centering
    \renewcommand{\arraystretch}{0.3}
    \setlength{\tabcolsep}{.5pt} %
    \large

    \newcommand{\sizew}{.16\linewidth}
    \newcommand{\sizeh}{.24\linewidth}

    \newcommand{\imgcell}[1]{%
        \includegraphics[width=\sizew]{images/realprev/#1_0.png}&%
         $\raisebox{5ex}{\Large$\boldsymbol{\rightarrow}$}$ &%
        \includegraphics[width=\sizew]{images/realprev/#1_1.png}
    }

    \newcommand{\spacersize}{.3ex}

    \resizebox{.99\linewidth}{!}{%
    \begin{tabular}{ccc S{\spacersize} ccc S{\spacersize} ccc S{\spacersize} ccc}
        \multicolumn{3}{c}{Beethoven} & \multicolumn{3}{c}{David} & \multicolumn{3}{c}{Schubert} & \multicolumn{3}{c}{Chopin}\\
        \cmidrule(lr){1-3}\cmidrule(lr){4-6}\cmidrule(lr){7-9}\cmidrule(lr){10-12}
        \imgcell{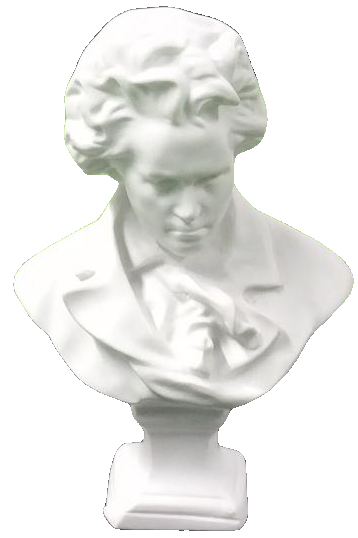} & \imgcell{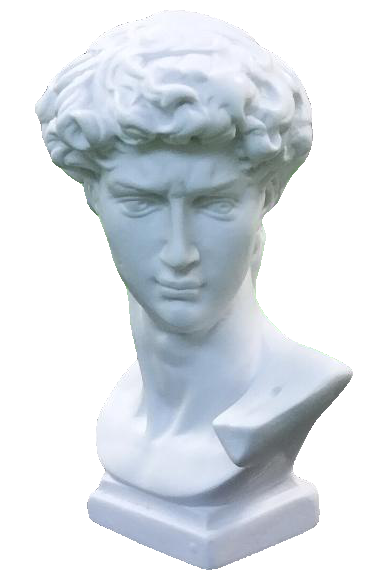} & \imgcell{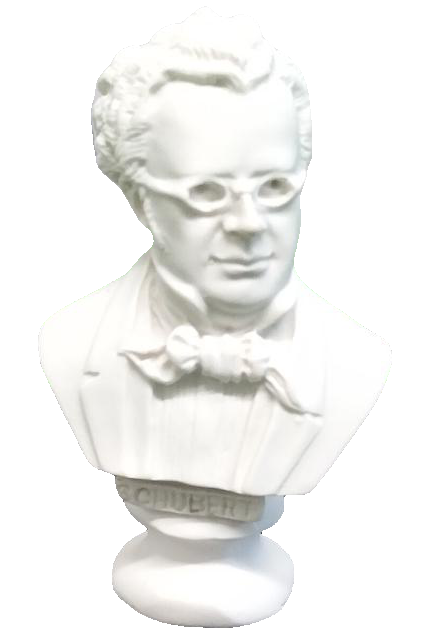} & \imgcell{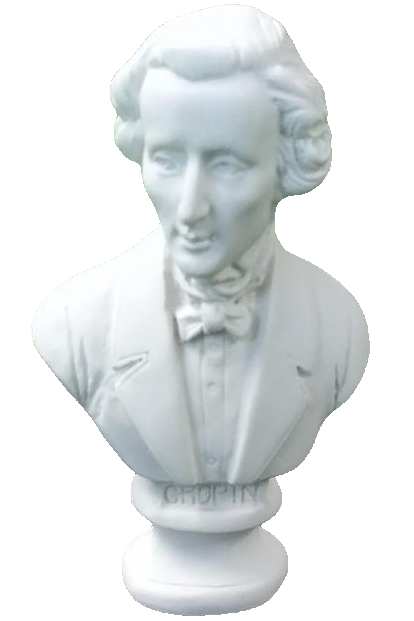}\\[2ex]

        \multicolumn{3}{c}{Wagner} & \multicolumn{3}{c}{Bach} & \multicolumn{3}{c}{Mozart} & \multicolumn{3}{c}{Muse}\\
        \cmidrule(lr){1-3}\cmidrule(lr){4-6}\cmidrule(lr){7-9}\cmidrule(lr){10-12}
        \imgcell{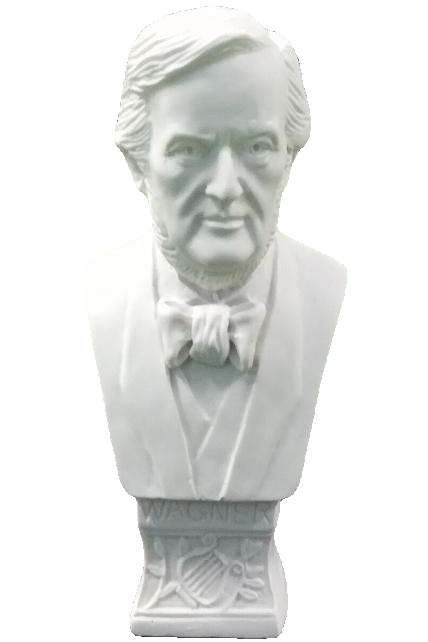} & \imgcell{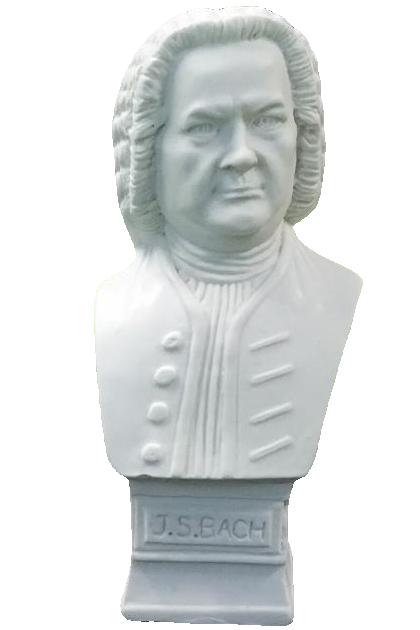} & \imgcell{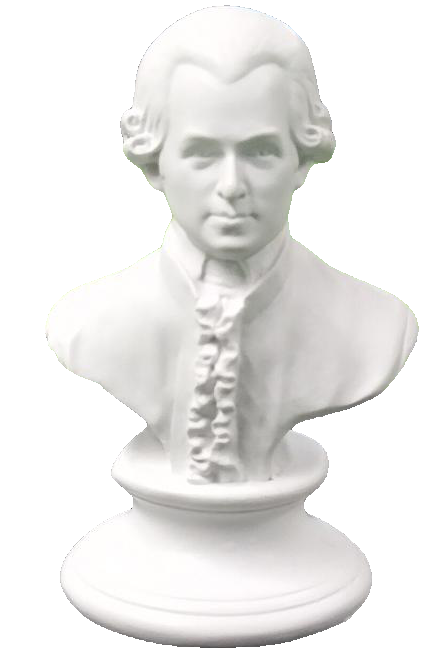} & \imgcell{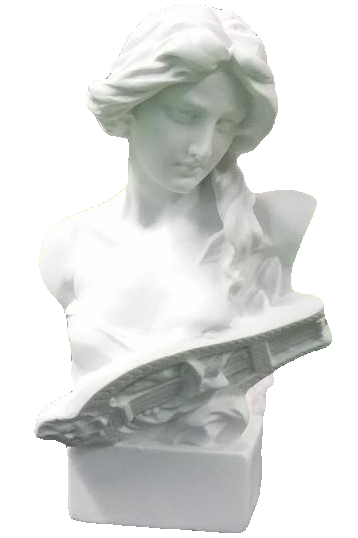}\\[2ex]

    \end{tabular}}

    \caption{\textbf{\RC{Real-world dataset}.} Different bust figurines were first photographed with and without various colored coats (Beethoven, David, Schubert, Chopin, Wagner, Bach) or glossy varnishes (Mozart, Muse).}
    \label{fig:real-dataset}
\end{figure}

\RC{Given the lack of available datasets suitable for studying BRDF transfer, we build two datasets: one synthetic with custom Blender shaders; and one real capturing figurines under varying material conditions. Both datasets will be shared publicly. Given the complexity of acquiring real-world ground truth decompositions, we follow the usual practice~\cite{jin2023tensoir} and report quantitative performance on the synthetic dataset only. Qualitative results are shown for both.}

\newparagraph{\RC{Synthetic dataset}.} \RC{We obtain} \datasetsize{} freely available and open-source 3D models compatible with the Blender PBR rendering pipeline. Models are processed individually and rescaled so they share a similar size. We design a simple set of shader transforms $\T{}\in\{\Tk{1},...,\Tk{n}\}$ in Blender to alter all materials in a given scene.
This allows us to control the global $\alpha$ ($0$: no change, $1$: fully transformed) of a scene and the type of transformation applied. Following the dataset creation procedure of NeRFactor \cite{Zhang2021nerfactor}, we render 100/20 training/test views for each scene. An overview of the dataset is provided in \cref{fig:syn-dataset}, which shows all scenes and synthetic transformations.
Next, we briefly describe the different synthetic transformations:
\begin{itemize}
    \item In `original', the original PBR materials are used;
    \item $\Tk{1}$: $\rough^\prime = 0$ and $\albedo^\prime$ has 30\% the HSV value of $\albedo$;
    \item $\Tk{2}$: $\rough^\prime = 0$ and $\albedo^\prime = 0.5\albedo + 0.5 \albedo_\text{red}$;
    \item $\Tk{3}$: $\rough^\prime = 1$ and $\albedo^\prime = 0.2\albedo + 0.8\albedo_\text{sand}$;
    \item $\Tk{4}$: $\rough^\prime = \rough$ and $\albedo^\prime$ has the opposite hue of $\albedo$.
\end{itemize}

\noindent\IL{where $\rho_{\text{sand}}$ and $\rho_{\text{red}}$ are two RGB colors chosen arbitrarily.} \RC{While realistic transformation would require complex shaders, we highlight that our choice of transformations is motivated by visual approximation of real-world transformations, being: wetness ($\Tk{1}$), fresh painting ($\Tk{2}$), dustiness ($\Tk{3}$), painting ($\Tk{4}$).}

\begin{figure}
    \centering
    \includegraphics[width=\linewidth]{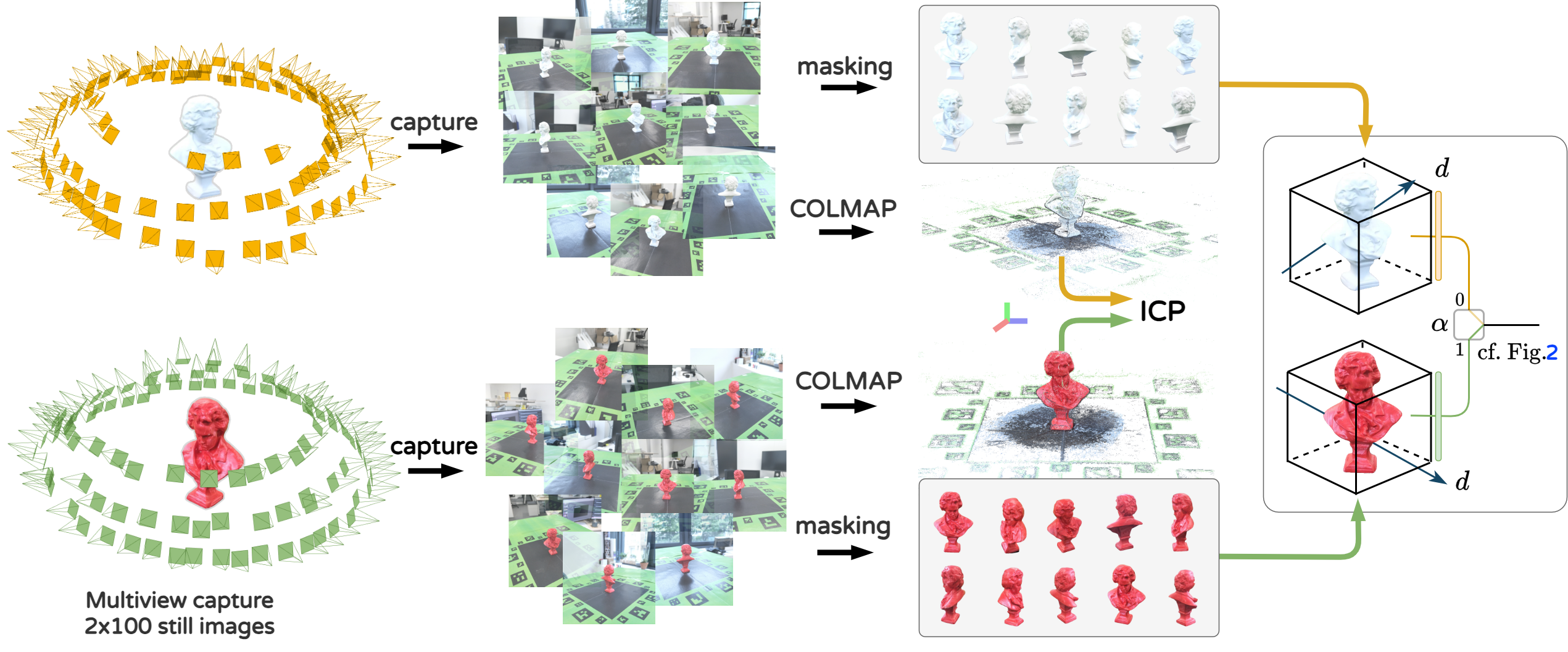}
    \caption{\textbf{Capture and pre-processing \IL{of} real data.} From left to right, we start by capturing still images for each variant of the object $\Szero$ (top) and $\Sone$ (bottom). Then we apply COLMAP to retrieve the dense reconstruction \RC{of both sets}. Using ICP, we align both shapes to set all camera poses on the \RC{same reference basis.}
    }
    \label{fig:real-pipeline}
\end{figure}

\newparagraph{\RC{Real-world dataset}.} \RC{We collected} \figurecount{} bust figurines (approximately 10 cm high, see \cref{fig:real-dataset}) and photographed them all around using a phone. We captured their appearance in two different conditions: first in their original appearance, and once more after \RC{altering their material condition, for example by} applying various colored coats or varnishes.
\JF{Unlike synthetic data, capturing real-world data results in unknown camera poses and two misaligned sets of photographs since they cannot be taken from exactly the same viewpoint. This can be prevented using a specially-designed camera rig as in \cite{toschi2023relight}, but comes at the cost and time of building and calibrating the apparatus.}
Instead, and as shown in \cref{fig:real-pipeline}, we apply COLMAP \cite{schoenberger2016sfm,schoenberger2016mvs} separately to the original and transformed sets of images and then estimate the rigid transformation matrix between the two resulting point clouds with the iterative closest point (ICP) algorithm. The resulting matrix is applied to correct the camera poses from both sets into a unique reference basis. \IL{Our optimization requires that we mask out the scene backgrounds, to do so we employ the library rembg~\cite{rembg} and manually correct the frames that are not masked properly.}

\newparagraph{Training.} We train in mixed batches with \IL{pixel rays from both $\Szero$ and~$\Sone$}.
All components are optimized end-to-end. The \IL{optimization of the target scene, on which we apply the learned transformation} is done without illumination embeddings $e_{\alpha}$. To apply the \IL{learned transform function on a new scene, we simply plug-in our trained MLP and compute $\Fi{}(\beta)$, shown on the right in \cref{fig:main}.}

\newparagraph{Network.} To implement our approach described in \cref{sec:method}, we adopt the same base architecture and optimization procedure as TensoIR~\cite{jin2023tensoir}. \IL{Vectors $e_\alpha$ are light embeddings of size $72$ that are used to} encode scene information that is specific to each $\alpha$ as both observations $\Szero$ and $\Sone$ might be captured under \IL{slightly} different lighting conditions. As in TensoIR, we use secondary ray marching to estimate the visibility mask $v_t$, \IL{but instead of sampling the radiance field, we leverage a dedicated $g_\text{indir}$ MLP to estimate the irradiance for occluded directions}. During the backward pass, the $\texttt{reduce-grad}$ function applies a $10^{-2}$ weight on the gradient \IL{in order to reduce its effect on directional inputs}. \IL{The model used to learn the transform function $\Fi{}$ is a small MLP with a single hidden layer of dimension $256$.}

\newparagraph{Baselines.}
To the best of our knowledge, there are no image-based material transform learning methods. Therefore, we made our best efforts to build strong baselines from existing techniques.

First, we note that TensoIR~\cite{jin2023tensoir} can be adapted by replacing the input of $\mathcal{D}_{\beta}$ from $\bar{a}_\alpha$ to $a_{\alpha}$ for material-specific information. This however poses two problems. First, $a_{\alpha}$ is not interpretable since it contains entangled information corresponding to the geometry, material, and illumination; second, transferring this function \IL{to a new scene would fail as the appearance features from both scenes would} belong to different embedding spaces.

\IL{We therefore choose a different approach to setup a fair baseline.} We first train on the original scene \IL{$s_0$ and the transformed scene $s_1$, separately}. Then, we extract the geometry and BRDF from both scenes by querying the volume. \IL{A MLP model is trained to learn the mapping between the two sets of BRDF.} Finally this model is applied on a new scene $s$, such as to map its material \IL{from $\beta$ to~$\F{}(\beta)$. We follow this for several methods: on the vanilla TensoIR itself} \cite{jin2023tensoir} as well as on two recent inverse rendering methods: NeRO~\cite{liu2023nero} and Relightable 3D Gaussians (RG3D)~\cite{R3DG2023}. \IL{Since some baselines predict an additional metalness component while we do not, we set their values to zero during optimization and novel view synthesis to avoid having an unfair advantage.}

\subsection{\RC{Main results}}
\label{sec:main-results}

\begin{table}[h]
	\centering
	\resizebox{.95\linewidth}{!}{%
		\begin{tabular}{c c c ccc cHH}
			\toprule
			\multirow{2}{*}{Transfer} &\multirow{2}{*}{Method} & Normals* & \multicolumn{3}{c}{Albedo} & \multicolumn{3}{c}{Render} \\
			\cmidrule(lr){3-3}\cmidrule(lr){4-6}\cmidrule(lr){7-9}
			&& MAE$_\downarrow$ & PSNR$_\uparrow$ & SSIM$_\uparrow$ & LPIPS$_\downarrow$ & PSNR$_\uparrow$ & SSIM$_\uparrow$ & LPIPS$_\downarrow$ \\
			\midrule

			\multirow{4}{*}{$\Tk{1}$}& NeRO & \underline{7.726} & 16.03 & 0.676 & 0.254 & 20.40 & 0.759 & 0.238 \\
			& TensoIR & 10.92 & \underline{17.82} & \underline{0.722} & 0.262 & 21.04 & 0.826 & 0.180 \\
			& R3DG & 11.41 & 11.57 & 0.610 & \underline{0.223} & \textbf{26.12} & \textbf{0.908} & \textbf{0.107} \\
			& \oc{}ours & \oc{}\textbf{6.750} & \oc{}\textbf{19.80} & \oc{}\textbf{0.781} & \oc{}\textbf{0.195} & \oc{}\underline{21.95} & \oc{}\underline{0.815} & \oc{}\underline{0.179} \\
			\midrule

			\multirow{4}{*}{$\Tk{2}$}& NeRO & \underline{7.726} & 16.54 & \underline{0.693} & 0.242 & 21.47 & 0.769 & 0.230 \\
			& TensoIR & 10.92 & \underline{16.66} & 0.688 & 0.251 & 20.02 & 0.815 & 0.170 \\
			& R3DG & 11.41 & 12.44 & 0.635 & \underline{0.219} & \textbf{27.36} & \textbf{0.918} & \textbf{0.101} \\
			& \oc{}ours & \oc{}\textbf{6.750} & \oc{}\textbf{18.62} & \oc{}\textbf{0.770} & \oc{}\textbf{0.199} & \oc{}\underline{22.44} & \oc{}0.832 & \oc{}0.171 \\
			\midrule

			\multirow{4}{*}{$\Tk{3}$}& NeRO & \underline{7.726} & 16.21 & 0.690 & 0.248 & 24.00 & 0.846 & 0.195 \\
			& TensoIR & 10.92 & \underline{16.90} & \underline{0.697} & 0.293 & 23.72 & \underline{0.890} & \underline{0.131} \\
			& R3DG & 11.41 & 12.79 & 0.656 & \underline{0.217} & \textbf{31.45} & \textbf{0.962} & \textbf{0.0633} \\
			& \oc{}ours &  \oc{}\textbf{6.750} & \oc{}\textbf{19.81} & \oc{}\textbf{0.787} & \oc{}\textbf{0.197} & \oc{}\underline{29.76} & \oc{}\underline{0.919} & \oc{}\underline{0.110} \\
			\bottomrule

			\multirow{4}{*}{$\Tk{4}$}& NeRO & \underline{7.726} & 15.43 & 0.662 & 0.263 & 22.60 & 0.807 & 0.219 \\
			& TensoIR & 10.92 & \underline{17.52} & \underline{0.696} & 0.273 & 22.61 & 0.884 & 0.133 \\
			& R3DG & 11.41 & 13.26 & 0.668 & \underline{0.217} & \textbf{29.12} & 0.954 & 0.0884 \\
			& \oc{}ours & \oc{}\textbf{6.750} & \oc{}\textbf{18.88} & \oc{}\textbf{0.766} & \oc{}\textbf{0.203} & \oc{}\underline{26.88} & \oc{}0.920 & \oc{}0.128 \\
			\bottomrule

			\multirow{4}{*}{mean}& NeRO & \underline{7.726} & 16.05 & 0.680 & 0.252 & 22.12 & 0.795 & 0.220 \\
			& TensoIR & 10.92 & \underline{17.23} & \underline{0.701} & 0.270 & 21.85 & 0.854 & 0.153 \\
			& R3DG & 11.41 & 12.52 & 0.642 & \underline{0.219} & \textbf{28.51} & 0.936 & 0.0901 \\
			& \oc{}ours & \oc{}\textbf{6.750} & \oc{}\textbf{19.28} & \oc{}\textbf{0.776} & \oc{}\textbf{0.199} & \oc{}\underline{25.26} & \oc{}0.872 & \oc{}0.147 \\
			\bottomrule

	\end{tabular}}\\
	{\footnotesize{} * Normals are independent of the transformation learned.}

	\caption{\textbf{Novel view transfer evaluation.} We evaluate the material estimation on our synthetic dataset by measuring metrics after transferring cross-scenes. Evaluation on the test set on novel view synthesis. We highlight \RC{\textbf{best} and \underline{2nd best}}.
	}
	\label{tab:transfer}
\end{table}
\begin{figure}[t]
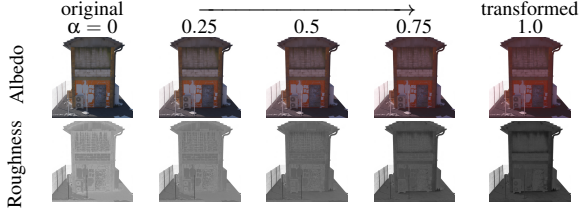

    \centering
    \scriptsize
    \newcommand{\imgsize}{.1\linewidth}
    \setlength{\tabcolsep}{.2pt} %
    \renewcommand{\arraystretch}{0.5} %

	\newcommand{\row}[2]{%
    \includegraphics[width=\imgsize]{images/interpolation/row#1_0_#2.png}&%
    \includegraphics[width=\imgsize]{images/interpolation/row#1_25_#2.png}&%
    \includegraphics[width=\imgsize]{images/interpolation/row#1_50_#2.png}&%
    \includegraphics[width=\imgsize]{images/interpolation/row#1_75_#2.png}&%
    \includegraphics[width=\imgsize]{images/interpolation/row#1_100_#2.png}%
    }

    \resizebox{.95\linewidth}{!}{%
    \tiny
    \setlength{\tabcolsep}{4pt} %
    \begin{tabular}{cccccc}
        &original & \multicolumn{3}{c}{$\xrightarrow{\hspace{10em}}$} & transformed\\
        &$\alpha=0$ & $0.25$ & $0.5$ & $0.75$ & $1.0$\\
        \verti{7ex}{Albedo} & \row{1}{albedo}\\
        \verti{8ex}{Roughness} & \row{1}{roughness}\\
    \end{tabular}}

    \caption{\textbf{Transform interpolation.} We linearly interpolate \IL{the BRDF parameters} between the original and target scene for varying values of $\alpha \in [0,1]$.}
    \label{fig:qual_alpha}
\end{figure}

\newparagraph{Learning BRDF transfer.} In \cref{tab:transfer} we present the BRDF transfer results averaged over all synthetic scenes for each transformation $\T{}$, and the mean over all transformations. To assess the quality of our decomposition and material transfer, we report Mean Angular Error (MAE) for normals~($\downarrow$) and PSNR~($\uparrow$), SSIM~($\uparrow$), and LPIPS~($\downarrow$) for the estimated albedo. For completeness, we also report PSNR~($\uparrow$) of the rendering for novel view synthesis.
Results in the table advocate that our method better estimates normals and albedo, showcasing that it faithfully learns the transformation of the BRDF function. Importantly, note that R3DG have higher fidelity renderings although this comes at the cost of inaccurate decomposition (normals, albedo) which suggests entangled material information in the scene lighting.
Corresponding qualitative results are shown in \cref{fig:qual_main1} and \cref{fig:qual_main2} for four scenes and three transformations. The latter demonstrates the superiority of our method, producing faithful renderings without compromising the scene decomposition (normals, albedo, roughness).
In \cref{fig:qual_alpha} we further show that our formulation allows interpolation between the original ($\alpha=0$) and the learned transformed BRDF ($\alpha=1$), according to \cref{eq:mapping}.

\begin{figure}[t]
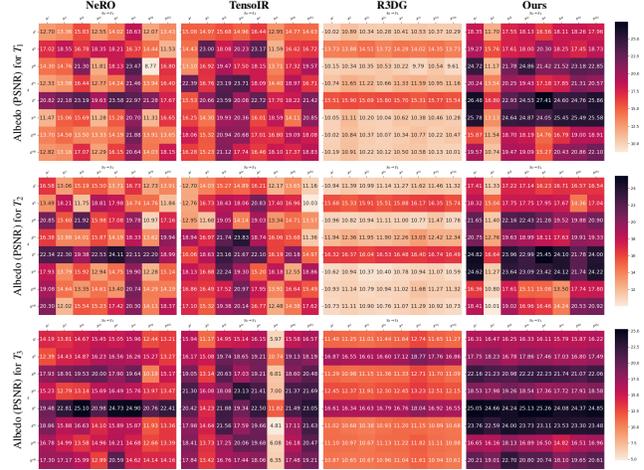

	\centering
	\scriptsize
	\newcommand{\colsize}{.13\linewidth}
	\setlength{\tabcolsep}{0pt} %
	\renewcommand{\arraystretch}{0} %
	\definecolor{mycolor}{HTML}{3FBC9D}

	\newcommand{\row}[3]{%
		\verti{24ex}{#1} &%
		\includegraphics[height=4.85cm, trim=0 .5cm 0 0, clip]{images/heatmaps/#2_#3_nero.png} &%
		\includegraphics[height=4.85cm, trim=0 .5cm 0 0, clip]{images/heatmaps/#2_#3_tensoir.png} &%
		\includegraphics[height=4.85cm, trim=0 .5cm 0 0, clip]{images/heatmaps/#2_#3_r3dg.png} &%
		\includegraphics[height=4.85cm, trim=0 .5cm 0 0, clip]{images/heatmaps/#2_#3_ours.png} &%
		\includegraphics[height=4.85cm, trim=0 .5cm 0 0, clip]{images/heatmaps/#2_#3_scale.png}%
	}

	\newcommand{\rowtrim}[3]{%
		\verti{28ex}{#1} &%
		\includegraphics[height=4.5cm, trim=0 0 0 2cm, clip]{images/heatmaps/#2_#3_nero.png} &%
		\includegraphics[height=4.5cm, trim=0 0 0 2cm, clip]{images/heatmaps/#2_#3_tensoir.png} &%
		\includegraphics[height=4.5cm, trim=0 0 0 2cm, clip]{images/heatmaps/#2_#3_r3dg.png} &%
		\includegraphics[height=4.5cm, trim=0 0 0 2cm, clip]{images/heatmaps/#2_#3_ours.png} &%
		\includegraphics[height=4.5cm, trim=0 0 0 2cm, clip]{images/heatmaps/#2_#3_scale.png}%
	}

	\resizebox{\linewidth}{!}{%
		\large
		\begin{tabular}{H C{2ex} ccccc}
			&& \textbf{NeRO} & \textbf{TensoIR} & \textbf{R3DG} & \textbf{Ours} \\

			\verti{15ex}{$T_1$} & \row{Albedo (PSNR) for $T_1$}{mushy}{PSNR_albedo}\\
			\verti{15ex}{$T_2$} & \row{Albedo (PSNR) for $T_2$}{varnish}{PSNR_albedo}\\

			\verti{15ex}{$T_3$} & \row{Albedo (PSNR) for $T_3$}{dust}{PSNR_albedo}\\
	\end{tabular}}

	\caption{\textbf{Performance of BRDF transfer per transformation.} We provide heatmaps of Albedo PSNR ($\uparrow$) for pairs of source (horizontal) and target (vertical) scenes. The diagonal indicates performance of the BRDF transformation when applied on the same scene. Of note, some scenes are easily transformed~(\IL{$\s{2}$, $\s{5}$}), arguably because of simpler appearance (cf. \cref{fig:syn-dataset}) while TensoIR seems to struggle to learn $T_3$ on the \IL{$\s{6}$} scene. Despite great rendering capability~(cf.~\cref{tab:transfer}), R3DG struggles to faithfully decompose the scene, while our method consistently outperforms all baselines.%
    }
	\label{fig:heatmap}
\end{figure}

Additionally, we provide heatmaps of per-scene Albedo performance in \cref{fig:heatmap} for all transformation, offering a more fine-grained analysis. Overall, heatmaps indicate that some scenes are easier to transform, such as \IL{$\s{2}$ and $\s{6}$} (cf.~\cref{fig:syn-dataset} for reference), whereas learning from \IL{$\s{6}$} proved to be more complex -- especially for TensoIR with transformation $T_3$. As shown in \cref{tab:transfer}, R3DG struggles to decompose the albedo correctly, while our method outperforms all baselines in almost all instances.

\begin{figure}
	\centering
	\footnotesize
	\renewcommand{\arraystretch}{0.2}
	\setlength{\tabcolsep}{0pt} %
	\newcommand{\imgW}{.166\linewidth}

	\newcommand{\row}[2]{%
		\includegraphics[width=\imgW]{images/i2i/c#1_input.png}&%
		\includegraphics[width=\imgW]{images/i2i/c#1.png}&%
		\includegraphics[width=\imgW]{images/i2i/c#1_gt.png}&%
		\includegraphics[width=\imgW]{images/i2i/c#2_input.png}&%
		\includegraphics[width=\imgW]{images/i2i/c#2.png}&%
		\includegraphics[width=\imgW]{images/i2i/c#2_gt.png}%
	}
	\resizebox{.99\linewidth}{!}{%
		\begin{tabular}{c S{.1ex} c S{.2ex} c S{.25ex} | S{.25ex} c S{.1ex} c S{.2ex} c}
			input & output & reference & input & output & reference \\
			\row{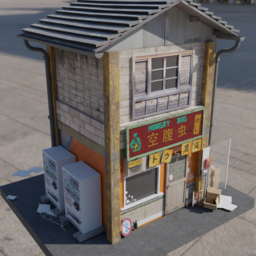}{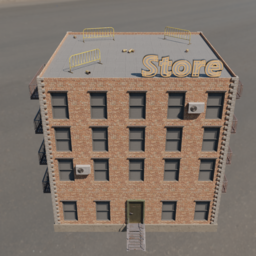}\\
			\row{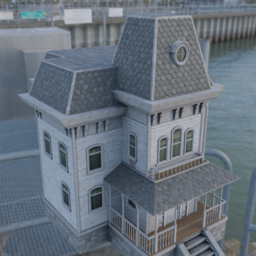}{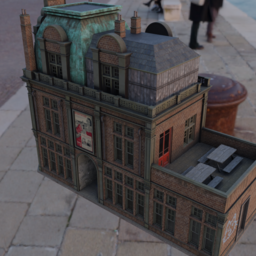}\\
	\end{tabular}}
	\caption{\textbf{Examples of prompt-driven translation.} InstructPix2Pix~\cite{brooks2022instructpix2pix} applied on images (\textit{with background}) with the input prompt ``make it more glossy''. Image translation methods limit the consistency of the output in terms of geometry and appearance. It also assumes the transformation is known and can be formulated as a prompt, \IL{which is not always the case}.
	}
	\label{fig:i2i}
\end{figure}

\newparagraph{Comparison to i2i translation methods.}  Additionally, we provide in \cref{fig:i2i} some results with the prompt guided image-to-image (i2i) translation method InstructPix2Pix \cite{brooks2022instructpix2pix}. The result exhibits the limitation of such techniques which require a priori knowledge about the transformation and its formulation with natural language. Also, the output is not geometrically consistent.


\subsection{Ablation study}

\RC{We conduct our ablations on the synthetic dataset of~\cref{sec:main-results} because of the availability of ground truth material properties. As before, we report metrics on estimated albedo and novel view synthesis.}
Here, we name our proposed method ``full model''; ``w/o \texttt{reduce-grad}'' allows the gradient to flow on the directional inputs of the illumination module without damping; ``w/o $g_{\text{dir}}, g_{\text{indir}}$'' removes the illumination MLPs and uses spherical gaussians and stratified sampling to represent light sources (as in TensoIR); ``w/o joint optim.'' corresponds to learning both scene representations separately and fitting a MLP to learn $\Fi{}{}$; ``w/o transfer'' acts as the lower bound with metrics of the original scene \IL{$\Szero$ computed against the reference images of $\Sone$}.

\newparagraph{Impact of design choices on the transfer capability.} In \cref{tab:ablation}, we present an evaluation of the components on the ability to transfer the learned transformation to other scenes. Using a neural light representation improves the geometry estimation \IL{as well as the overall quality of the predicted albedo. We note that learning on both the original and transformed scenes benefits the transfer of~$\Fi{}$}.

\begin{table}[h]
    \centering
    \renewcommand{\verti}[3]{\multirow{#3}{*}[#1]{\rotatebox[origin=c]{90}{#2}}}
    
    \resizebox{.93\linewidth}{!}{%
    \begin{tabular}{H l c ccc cHH}
        \toprule
        \multirow{2}{*}{Eval.}&\multirow{2}{*}{Ablations} & \multicolumn{1}{c}{Normals} & \multicolumn{3}{c}{Albedo} & \multicolumn{3}{c}{Render} \\
        \cmidrule(lr){3-3}\cmidrule(lr){4-6}\cmidrule(lr){7-9}
        && MAE$_\downarrow$ & PSNR$_\uparrow$ & SSIM$_\uparrow$ & LPIPS$_\downarrow$ & PSNR$_\uparrow$ & SSIM$_\uparrow$ & LPIPS$_\downarrow$ \\
         
        \midrule
        \multirow{3}{*}{$\Fi{i}(\scene{j})$}&%
        \oc{}ours (\textbf{full model}) & \oc{}\textbf{6.750} & \oc{}\textbf{19.80} & \oc{}\underline{0.781} & \oc{}\underline{0.195} & \oc{}\textbf{21.95} & \oc{}0.815 & \oc{}0.179 \\
        & \,$\drsh$ w/o $g_{\text{dir}}, g_{\text{indir}}$ & \underline{9.060} & \underline{18.51} & \textbf{0.784} & \textbf{0.187} & \underline{21.04} & \textbf{0.887} & 0.106 \\
        &\,\,$\drsh$ w/o joint optim. & 11.14 & 17.75 &	0.714 &	0.264 &	20.50 &	\underline{0.816} &	\underline{0.175} \\ %
        \bottomrule
    \end{tabular}}
    \caption{\textbf{\IL{Transfer} to other scenes.} Evaluates the transform $\T_{1}$, on novel view synthesis after applying \IL{$\Fi{}$ on a new scene $\Szero$. The resulting scene is evaluated against the reference images of $\Sone$, which were rendered with the ground truth transform $T_1$.}}
    \label{tab:ablation}
\end{table}

\newparagraph{Benefit of joint training.} It is also interesting to study the impact of each component when applying the learned BRDF \IL{transform} to the \emph{same scene} (rather than a new scene as done previously). 
This allows to evaluate the quality of the BRDF \IL{transform} learned on the source scene. 
Results from this experiment are presented in \cref{tab:ablation-self}. This shows the benefit of learning the transform function during optimization of the scene. We remark a slightly better MAE on ``w/o transfer'' as the geometry is learned on the original scene which is often easier to estimate compared to the transformed scene.

\begin{table}[h]
    \centering
    \resizebox{.90\linewidth}{!}{%
    \begin{tabular}{H l c ccc cHH}
        \toprule
        &\multirow{2}{*}{Ablations} & Normals & \multicolumn{3}{c}{Albedo} & \multicolumn{3}{c}{Render} \\
        \cmidrule(lr){3-3}\cmidrule(lr){4-6}\cmidrule(lr){7-9}
        && MAE$_\downarrow$ & PSNR$_\uparrow$ & SSIM$_\uparrow$ & LPIPS$_\downarrow$ & PSNR$_\uparrow$ & SSIM$_\uparrow$ & LPIPS$_\downarrow$ \\
    
        \midrule
        \multirow{4}{*}{$\Fi{i}(\scene{i}^*)$}&%
        \oc{}ours (\textbf{full model}) & \oc{}\underline{7.164} & \oc{}\underline{21.40} & \oc{}\underline{0.805} & \oc{}\underline{0.187} & \oc{}\textbf{30.51} & \oc{}\textbf{0.919} & \oc{}\underline{0.109} \\
        & \,$\drsh$ w/o $g_{\text{dir}}, g_{\text{indir}}$ & 9.338 & \textbf{23.80} & \textbf{0.851} & 0.210 & \underline{29.58} & \underline{0.900} & 0.143 \\
        &\,\,$\drsh$ w/o joint optim. & 11.14 & 19.43 &	0.747 & 0.242 & 22.37 & 0.833 & 0.159 \\ %
        & \,\,\,$\drsh$ w/o transfer & \textbf{6.750} & 18.56 & 0.769 & \textbf{0.175} & 21.12 & 0.888 & \textbf{0.106} \\
        \bottomrule
    \end{tabular}}
    \caption{\textbf{\IL{Transformation of} the same scene.} We measure the gain related to optimizing \IL{jointly $\Szero$ and $\Sone$} and ablate the different components. In ``w/o joint optim.'' the two scenes are optimized separately, while the last experiment ``w/o transfer'' \IL{is by evaluating straight from $\Szero$} with no transfer.}
    \label{tab:ablation-self}
\end{table}
\begin{figure}[h]
    \centering
    \footnotesize
    \setlength{\tabcolsep}{0pt} %
    \newcommand{\imgW}{.333\linewidth}

    \newcommand{\row}[1]{%
        \includegraphics[width=\imgW]{images/ablations/gt_#1.png}&%
        \includegraphics[width=\imgW]{images/ablations/ours_#1.png}&%
        \includegraphics[width=\imgW]{images/ablations/stopgrad_#1.png}&%
        \includegraphics[width=\imgW]{images/ablations/SGlight_#1.png}%
    }

    \resizebox{.99\linewidth}{!}{%
    \begin{tabular}{C{3ex} c S{.5ex} c S{.5ex} c S{.5ex} c}
         & GT & ours (\textbf{full model}) & w/o $\texttt{reduce-grad}$ & TensoIR \\

        \verti{8ex}{Albedo} & \row{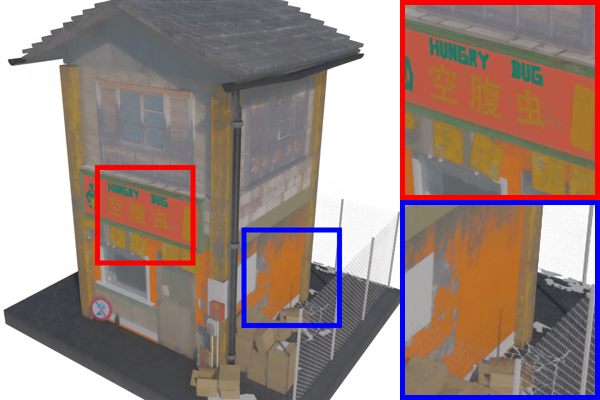} \\
        \verti{10ex}{Roughness} & \row{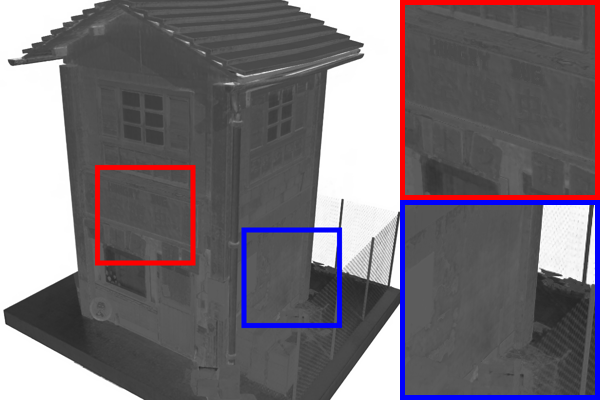} \\
        \verti{10ex}{Normals} & \row{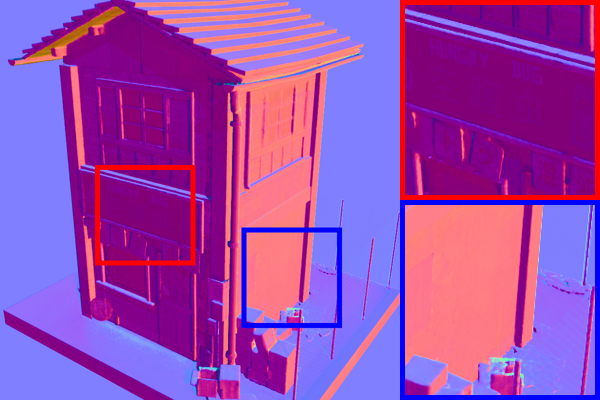} \\
        \verti{7ex}{$\ell_{\text{diff}}$} & \row{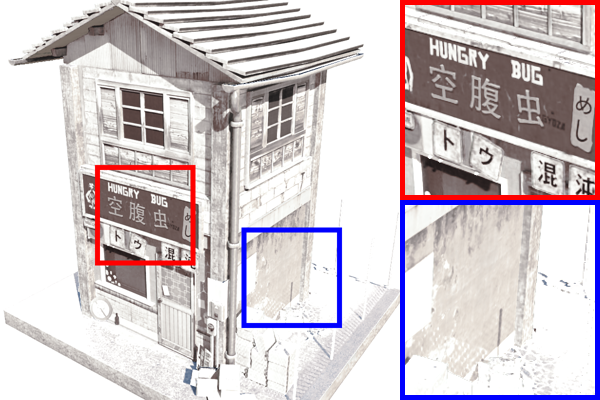} \\
        \verti{9ex}{Render} & \row{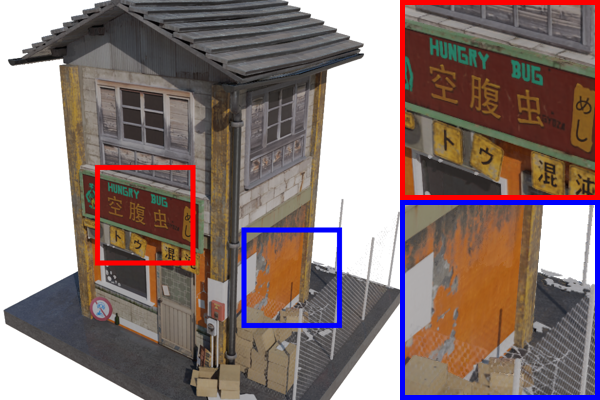} \\
    \end{tabular}}
    
    \caption{\textbf{Illumination ablation.} We show a detailed breakdown of the scene $\s{1}$, we notice that when learning from two scenes with a neural light representation, the diffuse light $\ell_{\text{diff}}$ tends to overfit to the geometry of the scene which leads to color information leaking from the albedo into the light. In contrast, reducing the gradient on directional input $n$ and $t$ of $g_{\text{dir}}$ alleviates this effect resulting in a uniform diffuse light.\vspace{-1em}}
    \label{fig:qual_ablation}
\end{figure}
\begin{figure}[h]
    \centering
    \footnotesize
    \renewcommand{\arraystretch}{0.2}
    \setlength{\tabcolsep}{0pt} %
    \newcommand{\imgW}{.333\linewidth}
    \newcommand{\imgH}{.166\linewidth}

    \newcommand{\row}[1]{%
        \includegraphics[width=\imgW]{images/toy/gt_#1.png}&%
        \includegraphics[width=\imgW]{images/toy/ours_#1.png}&%
        \includegraphics[width=\imgW]{images/toy/SGlight_#1.png}%
    }
    \resizebox{.99\linewidth}{!}{%
    \begin{tabular}{C{3ex} cc S{.5ex} cc S{.5ex} cc}
        & GT & ours (\textbf{full model}) &  TensoIR \\
        
        \verti{10ex}{Roughness} & \row{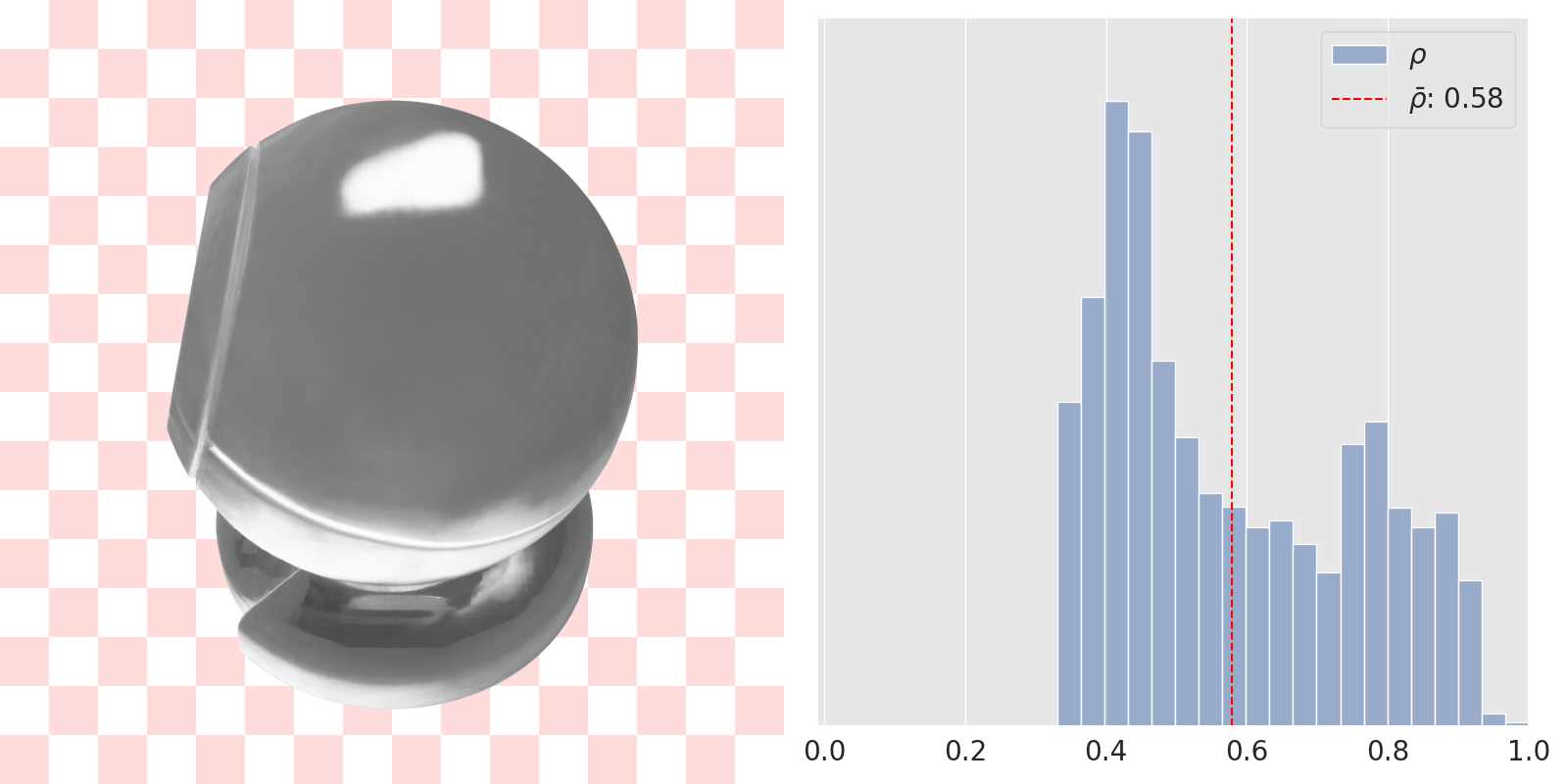} \\
        \verti{9ex}{Envmap} & \row{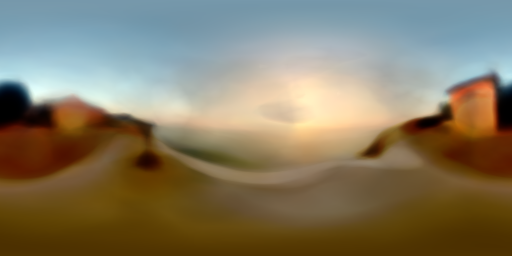} \\
        \verti{8ex}{$\ell_\text{spec}$} & \row{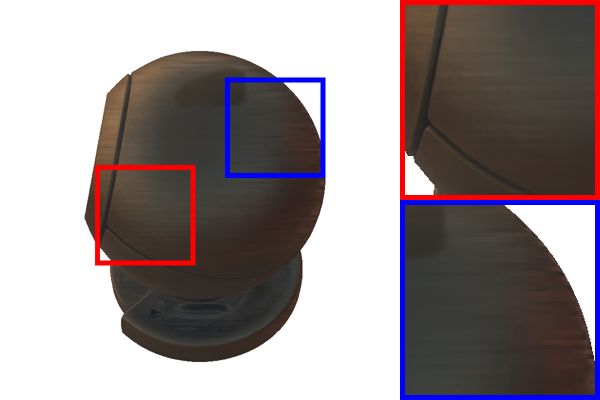} \\ %
        \verti{10ex}{Render} & \row{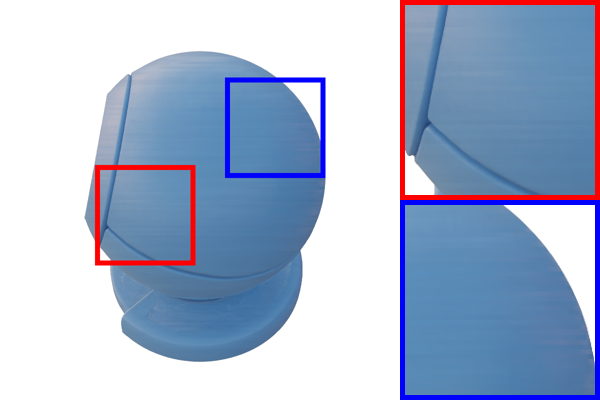} \\
        
        && 2.3h/optim, 16s/frame & 5.0h/optim, 30s/frame
        
    \end{tabular}}
    
    \caption{\textbf{Comparison to TensoIR,} which models illumination with spherical Gaussians and stratified sampling of the light directions. This model corresponds to w/o $g_{\text{dir}}, g_{\text{indir}}$ in our ablations. Instead, our method uses a neural representation to model pre-integrated illumination.}
    \label{fig:sglight}
\end{figure}

\newparagraph{Effect of \texttt{reduce-grad}.} When adopting integrated directional encoding (IDE) for the illumination components, we notice that the normals tend to degrade. 
Considering the shorter gradient path, it is much easier for the light MLP to bake albedo information at the expense of worsening surface normals and albedo. We can see this in the ``w/o \texttt{reduce-grad}'' column of \cref{fig:qual_ablation}: the billboard (cf. red zoom-in region) has high-frequency details baked into light while it is supposed to be a perfectly flat surface. Damping the gradient with the proposed \texttt{reduce-grad} operator prevents the normals from overfitting to the light gradient signal and results in a more uniform diffuse light estimation $\ell_\text{diff}$.

\newparagraph{Comparison to TensoIR.} The illumination used in TensoIR \cite{jin2023tensoir} revolves around stratified sampling which doesn't allow rendering low roughness surfaces. As such it is not capable of modeling reflective objects since all directions have the same probability of being sampled and there is no preference over the direction of reflection $t$. Spherical Gaussians do not allow for high-frequency details in the optimized environment map. Finally, using IDE provides an edge in terms of computation cost: on average, a scene optimization takes 2.3 hours, compared to 5.0 hours with TensoIR. We show in \cref{fig:sglight} a toy example with uniform glossy material \IL{along with a histogram of roughness values for each column (top). It shows that our model is able to capture higher frequency details while not requiring expensive and less accurate stratified sampling.}

\newparagraph{Ablations of the transfer network.} Further, we evaluated variations of our transfer network (\cref{sec:meth_mattransform}) for material mapping, increasing its capacity and adding residual connections. Our findings indicate that the network architecture has little effect on the performance, as we recorded less than 1.9\% difference in Normals MAE and 1.6\% in Albedo PSNR. This suggests that limited capacity is sufficient for learning BRDF a transformation.

\subsection{\RC{Real world transformations}}

\begin{figure}[t!]
    \centering
    \renewcommand{\arraystretch}{0.3}
    \setlength{\tabcolsep}{0pt} %
    \small

    \newcommand{\sizew}{.24\linewidth}
    \newcommand{\tensoirrow}[2]{%
        \multirow{2}{*}[7ex]{\includegraphics[height=\sizew,width=\sizew,trim=2cm 2cm 2cm 2cm, clip]{images/real/s0_#1_ref.png}}&%
        \multirow{2}{*}[0ex]{\raisebox{7ex}{$\rightarrow$}} &%
        \multirow{2}{*}[7ex]{\includegraphics[height=\sizew,width=\sizew,trim=2cm 2cm 2cm 2cm, clip]{images/real/s1_#1_ref.png}} &%
        \multirow{2}{*}[7ex]{\includegraphics[height=\sizew,width=\sizew,trim=2cm 2cm 2cm 2cm, clip]{images/real/s_#2_ref.png}} &%
        \verti{10ex}{TensoIR}&%
        \includegraphics[height=\sizew,width=\sizew,trim=2cm 2cm 2cm 2cm, clip]{images/real/tensoir_#2_normals.png}&%
        \includegraphics[height=\sizew,width=\sizew,trim=2cm 2cm 2cm 2cm, clip]{images/real/tensoir_#2-#1_albedo.png}&%
        \includegraphics[height=\sizew,width=\sizew,trim=2cm 2cm 2cm 2cm, clip]{images/real/tensoir_#2-#1_roughness.png}&%
        \includegraphics[height=\sizew,width=\sizew,trim=2cm 2cm 2cm 2cm, clip]{images/real/tensoir_#2-#1_render.png}&%

        \multirow{2}{*}[7ex]{\includegraphics[height=\sizew,width=\sizew,trim=2cm 2cm 2cm 2cm, clip]{images/real/gt_#2-#1_ref.png}}%
    }

    \newcommand{\oursrow}[2]{%
        &&&&%
        \verti{10ex}{ours}&%
        \includegraphics[height=\sizew,width=\sizew,trim=2cm 2cm 2cm 2cm, clip]{images/real/ours_#2_normals.png}&%
        \includegraphics[height=\sizew,width=\sizew,trim=2cm 2cm 2cm 2cm, clip]{images/real/ours_#2-#1_albedo.png}&%
        \includegraphics[height=\sizew,width=\sizew,trim=2cm 2cm 2cm 2cm, clip]{images/real/ours_#2-#1_roughness.png}&%
        \includegraphics[height=\sizew,width=\sizew,trim=2cm 2cm 2cm 2cm, clip]{images/real/ours_#2-#1_render.png}&%
    }

    \resizebox{.99\linewidth}{!}{%
    \begin{tabular}{cS{.2ex}cS{.2ex}c | c c cccc c}
        $\Szero$ && $\Sone$ & \textbf{Target} & & Normals & Albedo & Roughness & Render & Reference \\
        \tensoirrow{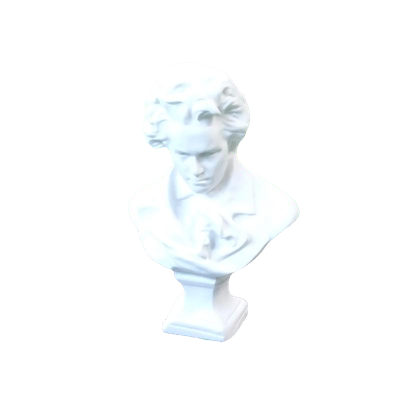}{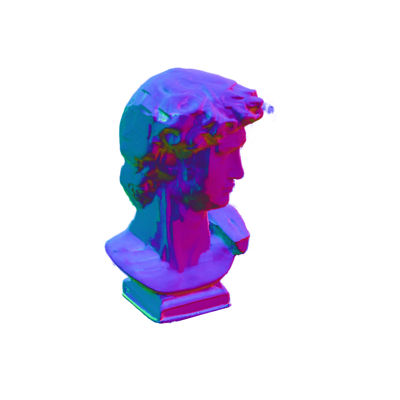} \\
        \oursrow{beethoven}{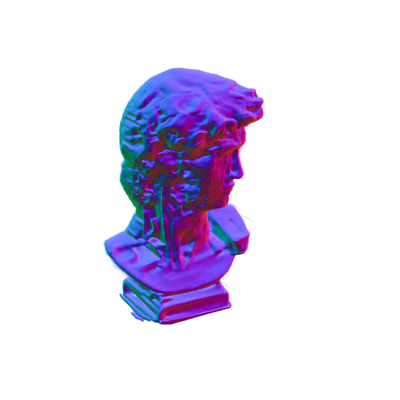} \\
        \tensoirrow{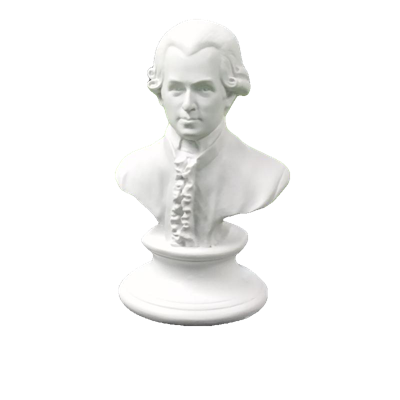}{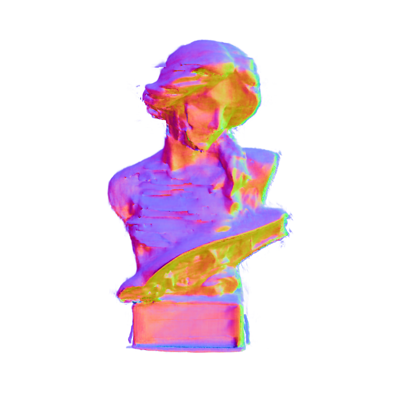} \\
        \oursrow{mozart}{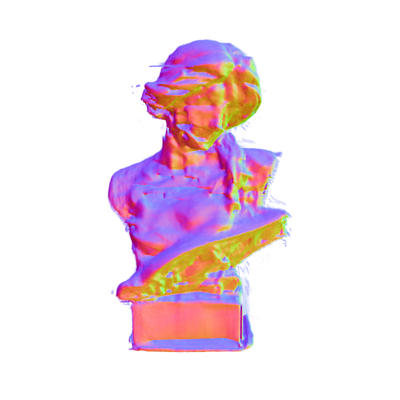} \\
        \tensoirrow{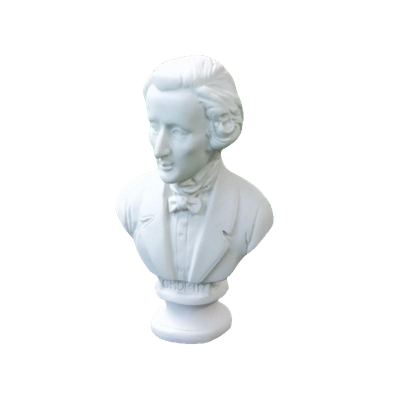}{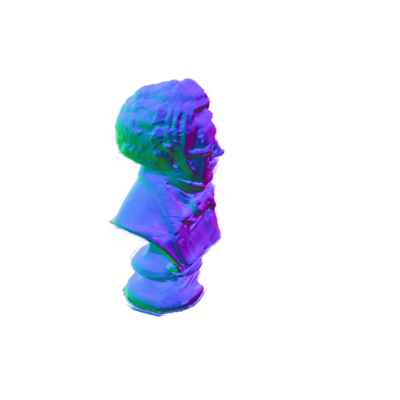} \\
        \oursrow{chopin}{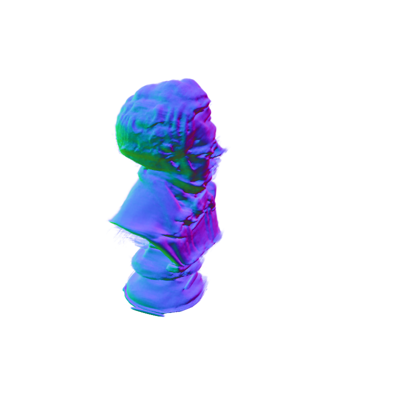} \\
        \tensoirrow{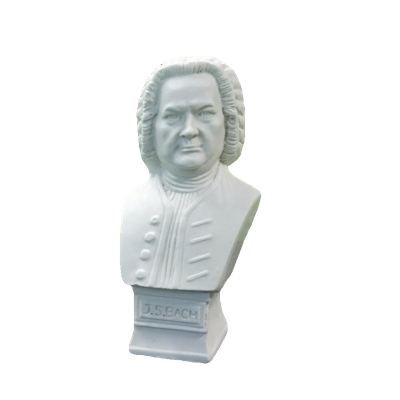}{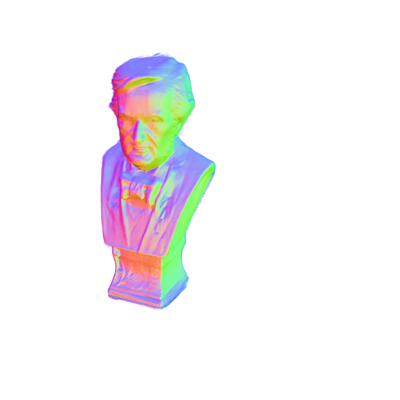} \\
        \oursrow{bach}{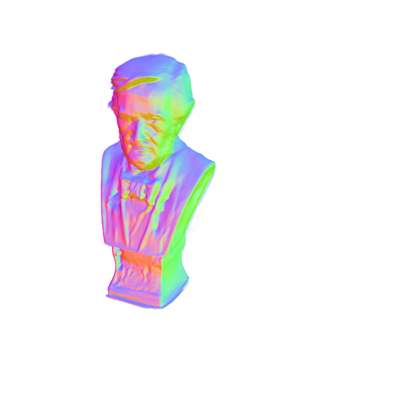} \\
    \end{tabular}}
    \caption{\textbf{Qualitative material transfers on real data.} We first learn the material transfer function from a figurine captured \IL{with} two different materials ($s_0$ and $s_1$, left). The learned transformation is then applied to a new figurine (target $s$, right), with the estimated normals, albedo, and roughness shown. Finally, the rendered object is compared to the reference photograph (far right). \IL{We provide results for TensoIR and our method.}\vspace{-1em}
    }
    \label{fig:real}
\end{figure}

In \cref{fig:real}, we qualitatively demonstrate the applicability of our method on our real-world figurines dataset. Compared to TensoIR, our decompositions are more accurate, particularly in terms of roughness, which is oversaturated by TensoIR (top two examples), and albedo, which TensoIR tends to darken (bottom two examples).
Altogether, this leads to our renderings being more realistic than TensoIR and more closing resembling the reference images. While we denote margin for improvement, we highlight that our method achieves believable results despite data being captured in relatively uncontrolled settings: handheld camera, possibly varying illumination conditions across captures, non-linear camera ISP, and errors in camera pose estimation. This demonstrates our method's effective robustness to these potential perturbations, showing that we are able to learn a material transfer from one figurine and apply it realistically to another.

{
\centering
\small
\newcommand{\imgsize}{.083\linewidth}
\newcommand{\imgsizez}{.1\linewidth}
\setlength{\tabcolsep}{0pt} %
\renewcommand{\arraystretch}{0.5} %

\newcommand{\row}[3]{%
    \includegraphics[width=\imgsize,height=\imgsize]{images/qual/#1_#2_normals.png}&%
    \includegraphics[width=\imgsize,height=\imgsize]{images/qual/#1_#2_albedo.png}&%
    \includegraphics[width=\imgsize,height=\imgsize]{images/qual/#1_#2_roughness.png}&%
    \includegraphics[width=\imgsize,height=\imgsize]{images/qual/#1_#2_render.png}&&%

    \includegraphics[width=\imgsize,height=\imgsize]{images/qual/#1_#2-#3_T1_albedo.png}&%
    \includegraphics[width=\imgsize,height=\imgsize]{images/qual/#1_#2-#3_T1_roughness.png}&%
    \includegraphics[width=\imgsize,height=\imgsize]{images/qual/#1_#2-#3_T1_render.png}&%

    \includegraphics[width=\imgsize,height=\imgsize]{images/qual/#1_#2-#3_T2_albedo.png}&%
    \includegraphics[width=\imgsize,height=\imgsize]{images/qual/#1_#2-#3_T2_roughness.png}&%
    \includegraphics[width=\imgsize,height=\imgsize]{images/qual/#1_#2-#3_T2_render.png}&%

    \includegraphics[width=\imgsize,height=\imgsize]{images/qual/#1_#2-#3_T3_albedo.png}&%
    \includegraphics[width=\imgsize,height=\imgsize]{images/qual/#1_#2-#3_T3_roughness.png}&%
    \includegraphics[width=\imgsize,height=\imgsize]{images/qual/#1_#2-#3_T3_render.png}%
}

\newcommand{\bcell}{\cellcolor{black!10}}
\newcommand{\bheader}{}
\newcommand{\rowheader}[2]{%
    \bheader{}
    \begin{tabular}{ccc}
        \\[-0.5ex]
        $\Szero$ & & $\Sone$ \\
        \includegraphics[width=\imgsizez,height=\imgsizez]{images/qual/si_#1_dry_render.png} &%
        $\raisebox{7.5ex}{\Large$\boldsymbol{\rightarrow}$}$ & \includegraphics[width=\imgsizez,height=\imgsizez]{images/qual/siprime_#1_#2_render.png}%
    \end{tabular}
}

\newcommand{\block}[2]{%
    && \multicolumn{4}{c}{} & \multicolumn{1}{c}{\verti{2.5ex}{\textbf{Source}}} & \multicolumn{3}{c}{\rowheader{#2}{T1}} & \multicolumn{3}{|c}{\rowheader{#2}{T2}} & \multicolumn{3}{|c}{\rowheader{#2}{T3}} \\
        \\
        && \multicolumn{4}{c}{\bcell{} \textbf{Target scene} $\beta$} &\multicolumn{1}{c}{}& \multicolumn{9}{c}{\bcell{} \textbf{Transformed target scene} $\mathcal{F}(\beta)$} \\
        \\[-.5ex]

        && Normals & Albedo & Roughness & Render && Albedo & Roughness & Render & Albedo & Roughness & Render & Albedo & Roughness & Render \\

        & \verti{7ex}{NeRO} & \row{nero}{#1}{#2}\\
        & \verti{8ex}{TensoIR} & \row{tensoir}{#1}{#2}\\
        & \verti{7ex}{R3DG} & \row{r3dg}{#1}{#2}\\
        & \verti{6ex}{Ours} & \row{ours}{#1}{#2}\\
        \multirow{-3}{*}{\verti{32ex}{}}&\verti{6ex}{\textbf{GT}} & \row{gt}{#1}{#2}\\
}

\begin{figure*}[]
    \resizebox{.99\linewidth}{!}{%
    \begin{tabular}{p{2ex} S{1ex} p{2.5ex} cccc p{2ex} ccc S{.6ex}|ccc S{.6ex}|ccc}
        \block{s2}{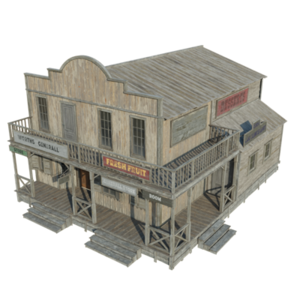} \\[10ex]
        \block{s1}{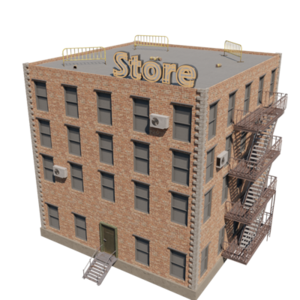}
    \end{tabular}}

    \caption{\textbf{Qualitative material transforms results.} We show qualitative results when synthesizing novel views with the learned transform function $\Fi{}$. \IL{For each sub-figure, we show in the top row the observed transform on the source scenes $(\Szero,\Sone)$, with three possible transformations: $T_1$, $T_2$, and $T_3$ column-wise. On the left, we show the optimization results of the target scene, and on the right, the transformed BRDF below the corresponding three source transforms.}
    }
    \label{fig:qual_main1}
\end{figure*}

\begin{figure*}[]
    \resizebox{.99\linewidth}{!}{%
    \begin{tabular}{p{2ex} S{1ex} p{2.5ex} cccc p{2ex} ccc S{.6ex}|ccc S{.6ex}|ccc}
        \block{s3}{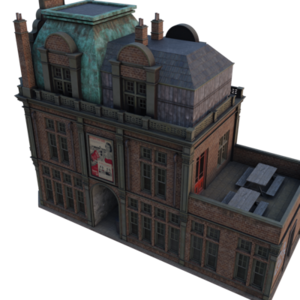} \\[10ex]
        \block{s4}{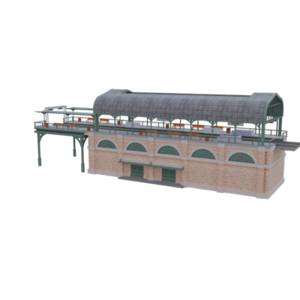}
    \end{tabular}}
    \caption{\textbf{Qualitative material transforms results.} We show qualitative results when synthesizing novel views with the learned transform function $\Fi{}$. \IL{For each sub-figure, we show in the top row the observed transform on the source scenes $(\Szero,\Sone)$, with three possible transformations: $T_1$, $T_2$, and $T_3$ column-wise. On the left, we show the optimization results of the target scene, and on the right, the transformed BRDF below the corresponding three source transforms.}
    }
    \label{fig:qual_main2}
\end{figure*}

}

\section{Limitations \& Conlusions}
\label{sec:conclusion}

We now discuss limitations and avenues for extending our method.

\newparagraph{Limitations.} Although the adopted pre-integrated neural light model is fast to optimize, it does not allow relighting the scene, as this would require retraining the model. Another limitation is that our method cannot handle hard-cast shadows; these shadows end up baked into the albedo, as is the case with all baselines. One solution could be to explicitly incorporate an occlusion estimation.
Additionally, our transformation is only applicable to materials that resemble those of the source scene. To expand the distribution domain of $\mathcal{F}$, it would be beneficial to learn the transformation from multiple scenes at once instead of just one.

\newparagraph{Problem setting.}
The current task is very underconstrained, estimating intrinsic parameters from multi-view inputs is already challenging in and of itself; here we aim to learn a BRDF mapping end to end from unaligned images. Not only the materials of $\Szero$ and $\Sone$ need to be correctly optimized, but the target scene should also be properly optimized since improper geometry estimation would inevitably lead to estimation of imprecise material transforms. 
For improved estimation of $\mathcal{F}$, a direction is to enforce a more controlled environment such as providing the scene geometry or imposing a fixed illumination.

\newparagraph{Extensions.}
An appealing avenue would be to work on richer material transformations. Currently, we assume the function to depend on the BRDF parameters alone and is point-wise, so there is no way to model spatially varying transformations. Furthermore, the transformation is uniform, \ie{} every point of the mesh with identical $\beta$ will result in a unique $\mathcal{F}(\beta)$. This is not always the case, for example for wetness, surfaces pointing upwards might be more affected than those pointing downwards. Introducing additional conditioning to the MLP modeling $\mathcal{F}$ or a spatially varying $\alpha$ could model that. Another interesting extension is to consider time-varying transformation such as in recent works by \cite{narumoto2024synthesizing}.

In this paper, we have introduced a challenging task of material transform estimation. Our proposed solution allows learning from two observations of the same scene with a single jointly optimized representation. The presented experiments demonstrate that the learned transformation can be transferred to new scenes. We hope this will motivate new research in this direction.

\newparagraph{Acknowledgments.}
This work was funded by the French Agence Nationale de la Recherche (ANR) with the project SIGHT (ANR-20-CE23-0016) and performed with HPC resources from GENCI-IDRIS (Grant  AD011014389R1). We thank Haian Jin for the useful discussion, and Mohammad Fahes and Anh-Quan Cao for their helpful comments.

\bibliographystyle{styles/eg-alpha-doi}
\bibliography{egbibsample}

\end{document}